\RequirePackage{fix-cm}
\documentclass[twocolumn]{svjour3}           % twocolumn
\usepackage{booktabs}
\usepackage{color}
\usepackage{graphicx}
\usepackage[pagebackref=true,colorlinks,citecolor=blue,linkcolor=red]{hyperref}
\usepackage{multirow}
\usepackage{natbib}
\usepackage{tabularx}
\usepackage{url}

\usepackage{tikz}
\usepackage{pgf, pgfsys, pgffor}
\usepackage{pgfplots}

\newcolumntype{Y}{>{\centering\arraybackslash}X}
\journalname{International Journal of Computer Vision}
\begin{document}

\title{Pix2Vox++: Multi-scale Context-aware 3D Object Reconstruction from Single and Multiple Images}
%\titlerunning{Short form of title}

\author{Haozhe Xie$^{1,2}$ \and
        Hongxun Yao$^{1}$ \and
        Shengping Zhang$^{1,4}$ \and \\
        Shangchen Zhou$^{3}$ \and
        Wenxiu Sun$^{2}$
}
\authorrunning{Haozhe Xie {\it et al.}}

\institute{%
Haozhe Xie\\ \email{hzxie@hit.edu.cn} \\~\\
Hongxun Yao\\ \email{h.yao@hit.edu.cn} \\~\\
Shengping Zhang\\ \email{s.zhang@hit.edu.cn} \\~\\
Shangchen Zhou\\ \email{shangchenzhou@gmail.com} \\~\\
Wenxiu Sun\\ \email{sunwenxiu@sensetime.com} \\~\\
$^1$ Harbin Institute of Technology, China \\
$^2$ SenseTime Research, China \\
$^3$ Nanyang Technological University, Singapore \\
$^4$ Peng Cheng Laboratory, China
}

\date{%
\textbf{Received:} 24 December 2019 /
\textbf{Revised:}  28 March 2020 / 
\textbf{Accepted:} 12 June 2020}

% \linenumbers
\maketitle

\begin{abstract}
Recovering the 3D shape of an object from single or multiple images with deep neural networks has been attracting increasing attention in the past few years.
Mainstream works ({\it e.g.} 3D-R2N2) use recurrent neural networks (RNNs) to sequentially fuse feature maps of input images.
However, RNN-based approaches are unable to produce consistent reconstruction results when given the same input images with different orders.
Moreover, RNNs may forget important features from early input images due to long-term memory loss.
To address these issues, we propose a novel framework for single-view and multi-view 3D object reconstruction, named Pix2Vox++.
By using a well-designed encoder-decoder, it generates a coarse 3D volume from each input image.
A multi-scale context-aware fusion module is then introduced to adaptively select high-quality reconstructions for different parts from all coarse 3D volumes to obtain a fused 3D volume.
To further correct the wrongly recovered parts in the fused 3D volume, a refiner is adopted to generate the final output.
Experimental results on the ShapeNet, Pix3D, and Things3D benchmarks show that Pix2Vox++ performs favorably against state-of-the-art methods in terms of both accuracy and efficiency.
\keywords{%
3D object reconstruction \and
Multi-scale \and 
Context-aware \and
Convolutional neural network}
\end{abstract}

\section{Introduction}

\begin{figure*}[!t]
  \centering
  \resizebox{\linewidth}{!} {
    \includegraphics{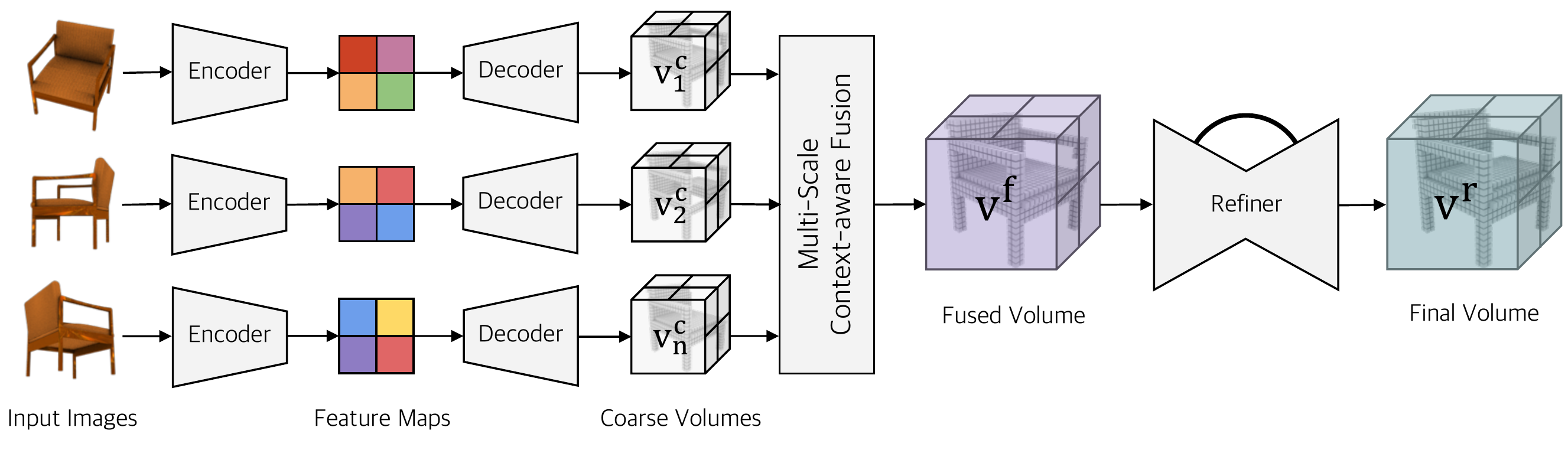}
  }
  \caption{Overview of the proposed Pix2Vox++. The network recovers the 3D shape of an object  from arbitrary (uncalibrated) single or multiple images. The reconstruction result can be refined when more input images are available. Note that the weights of the encoder and decoder are shared among all views.}
  \label{fig:overview}
\end{figure*}

Inferring the complete and precise 3D shape of an object is essential in robotics, 3D modeling and animation, object recognition, and medical diagnosis.
Traditional methods, such as Structure from Motion (SfM) \citep{DBLP:journals/acta/OnurVRA17} and Simultaneous Localization and Mapping (SLAM) \citep{DBLP:journals/air/Fuentes-PachecoAR15}, match features across images captured from slightly different views, and then use the triangulation principle to recover 3D coordinates of the image pixels.
Although these methods can produce 3D reconstruction with satisfactory quality, they typically capture multiple images of the same object using well-calibrated cameras, which is not practical or feasible in some situations \citep{DBLP:journals/pami/YangRMTW19}.

Recently, several deep learning-based approaches, including 3D-R2N2 \citep{DBLP:conf/eccv/ChoyXGCS16}, LSM \citep{DBLP:conf/nips/KarHM17}, DeepMVS \citep{DBLP:conf/cvpr/HuangMKAH18}, RayNet \citep{DBLP:conf/cvpr/PaschalidouUSGG18}, and AttSets \citep{DBLP:journals/ijcv/YangSAN19}, recover the 3D shape of an object from one or more RGB images without complex camera calibration and show promising results.
Both 3D-R2N2 \citep{DBLP:conf/eccv/ChoyXGCS16} and LSM \citep{DBLP:conf/nips/KarHM17} formulate multi-view 3D reconstruction as a sequence learning problem and fuse multiple feature maps extracted by a shared encoder from input images using recurrent neural networks (RNNs).
The feature maps are incrementally refined when more views of the object are available.
However, RNN-based methods suffer from three limitations.
First, RNNs are unable to consistently estimate the 3D shape of an object due to permutation variance \citep{DBLP:conf/iclr/VinyalsBK16} when given the same images with different orders.
Second, important features of early input images may be forgotten \citep{DBLP:conf/icml/PascanuMB13} due to long-term memory loss in RNNs.
Third, RNN-based methods are time-consuming since input images are processed sequentially without parallelization \citep{DBLP:conf/icassp/HwangS15}.

To overcome the shortcomings of the RNN-based methods, DeepMVS \citep{DBLP:conf/cvpr/HuangMKAH18} employs max pooling to aggregate deep features across a set of unordered images for multi-view stereo reconstruction.
RayNet \citep{DBLP:conf/cvpr/PaschalidouUSGG18} applies average pooling to aggregate the deep features extracted from the same voxel to recover the 3D structure.
Both max and average pooling eliminate the above limitations of RNNs,
but they capture only max or mean values without learning to attentively preserve useful information.
Very recent AttSets \citep{DBLP:journals/ijcv/YangSAN19} uses an attentional aggregation module to automatically predict a weight matrix as attention scores for input features.
However, aggregating features before the decoder is challenging for images with complex backgrounds and may cause problems in reconstructing objects in real-world scenarios.

To address the issues mentioned above, we propose Pix2Vox++, a novel framework for single-view and multi-view 3D reconstruction.
It contains four modules: encoder, multi-scale context-aware fusion, decoder, and refiner.
As shown in Figure \ref{fig:overview}, the encoder and decoder generate coarse 3D volumes from multiple input images in parallel, which eliminate the effect of orders of input images and accelerate computation.
The multi-scale context-aware fusion module then selects high-quality reconstructions from all coarse 3D volumes and generates a fused 3D volume, which exploits information from all input images without long-term memory loss.
Finally, the refiner further corrects the wrongly recovered parts of the fused 3D volumes to obtain a refined reconstruction.

The contributions can be summarized as follows:

\begin{itemize}
  \item We present a unified framework for both single-view and multi-view 3D object reconstruction, namely Pix2Vox++. It is composed of a well-designed encoder, decoder, and refiner, which shows strong abilities to handle 3D reconstruction in both synthetic and real-world images.
  \item We propose a multi-scale context-aware fusion module to adaptively select high-quality reconstructions for each part from different coarse 3D volumes in parallel to produce a fused reconstruction of the whole object.
  \item We construct a large-scale dataset, named {\it Things3D}, containing 1.68M images of 280K objects collected from over 39K indoor scenarios. To the best of our knowledge, it is the first large-scale dataset for multi-view 3D object reconstruction from naturalistic images.
  \item Experimental results on the ShapeNet, Pix3D, and Things3D datasets demonstrate that the proposed approaches outperform state-of-the-art methods in terms of both accuracy and efficiency.
\end{itemize}

A preliminary version of this work has been published in ICCV 2019 \citep{DBLP:conf/iccv/XieHXSS19}.
We make several extensions in this work compared to the preliminary version.
First, we replace VGG \citep{DBLP:conf/iclr/SimonyanZ14a} with ResNet \citep{DBLP:conf/cvpr/HeZRS16} as the new backbone network.
The improved method contains 25\% fewer parameters and is 5\% faster during inference compared to the one in the preliminary version.
In addition, there is a 1.5\% increase in Intersection-over-Union (IoU) on ShapeNet.
Second, we propose the multi-scale context-aware fusion to aggregate multi-scale features from multiple coarse 3D volumes.
Compared to the context-aware fusion in the preliminary version, it brings about a 1\% increase in IoU for multi-view reconstruction at $128^3$ resolution.
Third, we add several layers in the decoder to generate 3D volumes with higher resolutions of $64^3$ and $128^3$ which preserve better details of 3D objects.
Finally, we propose a large-scale naturalistic dataset for multi-view 3D reconstruction, which provides 708 times more 3D models than Pix3D \citep{DBLP:conf/cvpr/Sun0ZZZXTF18}.
Codes and pretrained models are publicly available at 
\url{https://gitlab.com/hzxie/Pix2Vox}.

\section{Related Work}

\begin{figure*}
  \centering
  \resizebox{\linewidth}{!} {
    \includegraphics{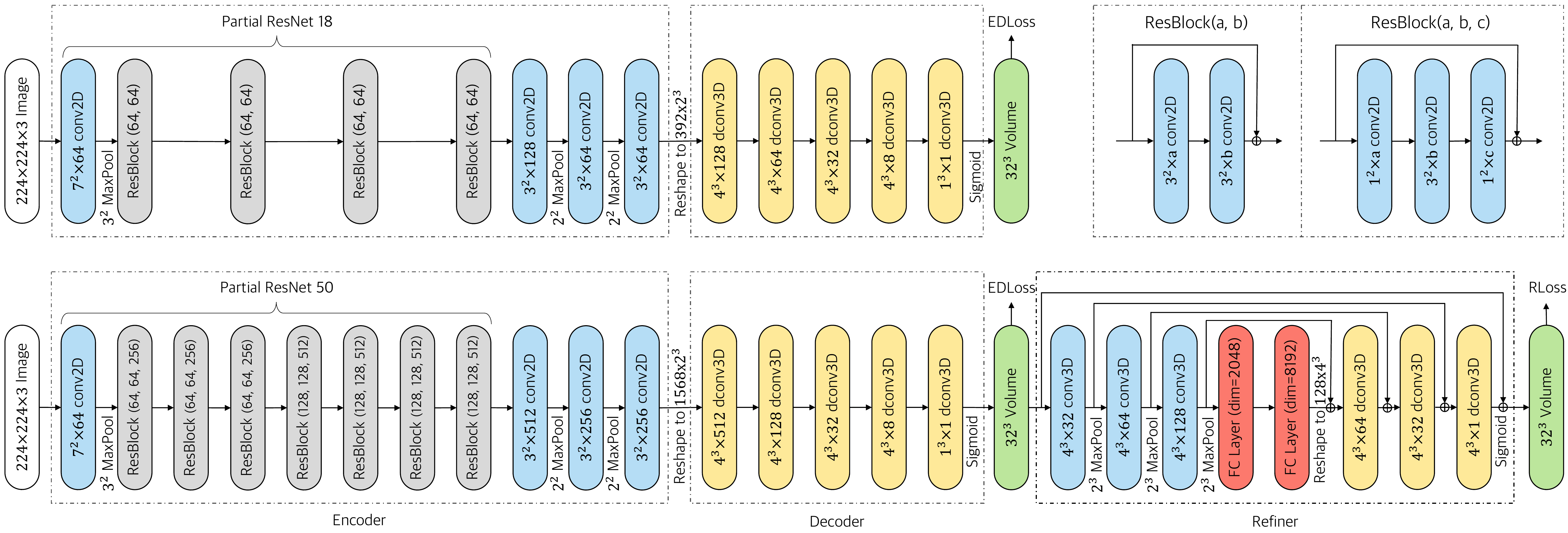}
  }
  \caption{The network architectures of Pix2Vox++/F (top) and Pix2Vox++/A (bottom) for low-resolution reconstruction. EDLoss and RLoss are defined in Equation \ref{eq:bce-loss}. To reduce the model size, the refiner is removed in Pix2Vox++/F.}
  \label{fig:network-architecture}
\end{figure*}

In this section, we review 3D object reconstruction methods closely related to this work.
Comprehensive reviews of 3D object reconstruction approaches can be found in \cite{DBLP:journals/corr/abs-1906-06543}.

\noindent \textbf{Single-view 3D Reconstruction.}
Predicting the complete 3D shape of an object from a single image is a long-standing and extremely challenging task.
Many attempts have been made to address this issue, such as Shape from X \citep{DBLP:journals/pami/BarronM15}, where X may represent silhouettes \citep{DBLP:conf/cvpr/DibraJOZG17}, shading \citep{DBLP:conf/cvpr/RichterR15}, or texture \citep{DBLP:journals/ai/Witkin81}.
However, these methods are rarely applied to real-world scenarios, as they all require strong presumptions and abundant expertise in natural images \citep{DBLP:journals/access/ZhangLLPL19}.
With the success of generative adversarial networks (GANs) \citep{DBLP:conf/nips/GoodfellowPMXWOCB14} and variational autoencoders (VAEs) \citep{DBLP:conf/iclr/KingmaW14}, 3D-VAE-GAN \citep{DBLP:conf/nips/0001ZXFT16} uses GAN and VAE to generate 3D reconstructions by taking a single-view image as input.
MarrNet \citep{DBLP:conf/nips/0001WXSFT17} reconstructs 3D objects by estimating depth, surface normals, and silhouettes of 2D images.
ShapeHD \citep{DBLP:conf/eccv/WuZZZFT18} extends MarrNet by incorporating a shape naturalness network to improve reconstruction results.
Kato and Harada \citep{DBLP:conf/cvpr/KatoH19} adopt a discriminator to ensure that reconstructed shapes are reasonable from any viewpoint.
OGN \citep{DBLP:conf/iccv/TatarchenkoDB17} uses octree to represent high-resolution 3D volumes with a limited memory budget.
Matryoshka Networks \citep{DBLP:conf/cvpr/Richter018} recursively decomposes a 3D shape into nested shape layers, which outperforms octree-based reconstruction methods.
Recently, several representations for 3D models, including point cloud \citep{DBLP:conf/cvpr/FanSG17}, mesh \citep{DBLP:conf/eccv/WangZLFLJ18}, and signed distance field \citep{DBLP:conf/nips/XuWDRU19}, have been adopted in 3D object reconstruction to reduce memory requirements for high-resolution 3D volumetric grids.
PSG \citep{DBLP:conf/cvpr/FanSG17} firstly recovers a point cloud from a single image with deep neural networks.
Pixel2Mesh \citep{DBLP:conf/eccv/WangZLFLJ18} is the first to reconstruct the 3D shape in a triangular mesh from a single image.
DISN \citep{DBLP:conf/nips/XuWDRU19} predicts the underlying signed distance field given a single input image.
With available fine-grained 3D part annotations \citep{DBLP:conf/cvpr/MoZCYTGS19}, several methods \citep{DBLP:conf/cvpr/PaschalidouGG20,DBLP:journals/tog/MoGYSWMG19,DBLP:journals/tog/ZhuXCYZ18} perform 3D reconstruction by compositing 3D parts in a hierarchical manner.
However, single-view 3D object reconstruction is an ill-posed and inherently ambiguous problem since partial observation of an object can theoretically be associated with an infinite number of possible 3D models.

\noindent \textbf{Multi-view 3D Reconstruction.}
Traditionally, 3D dense reconstruction in SfM and SLAM requires a collection of RGB images \citep{DBLP:books/daglib/0015576}.
The 3D structure of an object is recovered by dense feature extraction and matching, or by minimizing reprojection errors \citep{DBLP:journals/trob/CadenaCCLSN0L16}.
However, the matching process becomes extremely difficult when multiple viewpoints are separated by a large margin.
Furthermore, scanning all surfaces of an object before reconstruction is sometimes impossible, leading to incomplete 3D shapes with occluded or hollowed-out areas \citep{DBLP:journals/pami/YangRMTW19}.
Recently, deep neural networks have been designed to learn the 3D shape from multiple RGB images.
Both 3D-R2N2 \citep{DBLP:conf/eccv/ChoyXGCS16} and LSM \citep{DBLP:conf/nips/KarHM17} are RNN-based, resulting in the networks being permutation variant and inefficient for aggregating features from long sequence images.
DeepMVS \citep{DBLP:conf/cvpr/HuangMKAH18} and RayNet \citep{DBLP:conf/cvpr/PaschalidouUSGG18} employ max and average pooling to aggregate deep features.
Recent AttSets \citep{DBLP:journals/ijcv/YangSAN19} uses an attentional aggregation module to effectively aggregate deep features.
However, these methods capture only partial information, ignoring many deep features, which may lead to low-quality reconstruction.
In addition to volumetric representations, recent works also reconstruct  3D objects in the form of point clouds and meshes.
\cite{DBLP:conf/aaai/LinKL18} use 2D convolutional operations to predict a dense point cloud from multiple viewpoints and jointly apply geometric reasoning with 2D projection optimization.
Pixel2Mesh++ \citep{DBLP:conf/iccv/WenZLF19} recovers 3D mesh by leveraging cross-view information with a graph convolutional network.
\cite{DBLP:conf/cvpr/LinWRSKFL19} reconstruct 3D objects from aligned videos by optimizing object meshes for multi-view photometric consistency while constraining mesh deformations with a shape prior.
These methods require extrinsic camera parameters or aligned images as input.
However, it is not always feasible to obtain extrinsic camera parameters, especially from those scenarios that viewpoints are separated by a large margin.

\section{The Proposed Method: Pix2Vox++}

\subsection{Overview}

The proposed Pix2Vox++ aims to reconstruct the 3D shape of an object from either single or multiple RGB images. 
The 3D shape of an object is represented by a 3D voxel grid, where $0$ and $1$ denote an empty cell and an occupied cell, respectively.
The key components of Pix2Vox++ are illustrated in Figure \ref{fig:overview}.
First, the encoder produces feature maps from input images.
Second, the decoder takes each feature map as input and correspondingly generates a coarse 3D volume.
Third, single or multiple 3D volumes are forwarded to the multi-scale context-aware fusion module that adaptively selects high-quality reconstructions for different parts from all coarse 3D volumes in parallel and generates a fused 3D volume.
Finally, the refiner further corrects the wrongly recovered parts of the fused 3D volume to produce the final reconstruction result.

\subsection{Network Architecture}

To achieve a good balance between accuracy and model size, we implement two versions of the proposed framework: Pix2Vox++/F and Pix2Vox++/A, as shown in Figure \ref{fig:network-architecture}.
% where F and A are short for fast and accurate, respectively.
The former involves much fewer parameters and lower computational complexity.
The latter has more parameters, which can reconstruct more accurate 3D shapes but has higher computational complexity.

\subsubsection{Encoder}

The encoder aims to compute a set of features for the decoder to recover the 3D shape of the object.
The first three convolutional blocks of ResNet \citep{DBLP:conf/cvpr/HeZRS16} are used to obtain a $512 \times 28^2$ feature map from a $224 \times 224 \times 3$ image.
We adopt ResNet-18 and ResNet-50 for Pix2Vox++/F and Pix2Vox++/A, respectively.
ResNet is followed by three sets of 2D convolutional layers, batch normalization layers, and ReLU layers to embed semantic information into feature maps.
The kernel sizes of the three convolutional layers are $3^2$, with padding of $1$.
There is a max pooling layer with a kernel size of $2^2$ after the second and third ReLU layers.
In Pix2Vox++/F, the output channels in the convolutional layer are numbered $128$, $64$ and $64$, respectively.
In Pix2Vox++/A, the output channels of the three convolutional layers are $512$, $256$ and $256$, respectively.
The feature maps produced by Pix2Vox++/F and Pix2Vox++/A are of sizes $64 \times 7^2$ and $256 \times 7^2$, respectively.

\subsubsection{Decoder}

The decoder is responsible for transforming information of 2D feature maps into 3D volumes.

\noindent \textbf{Low-resolution Reconstruction.}
There are five 3D transposed convolutional layers in both Pix2Vox++/F and Pix2Vox++/A.
Specifically, the first four transposed convolutional layers are of kernel sizes $4^3$, with strides of $2$ and paddings of $1$.
There is an additional transposed convolutional layer with a bank of $1^3$ filter.
Each transposed convolutional layer is followed by a batch normalization layer and a ReLU activation except for the last layer followed by a sigmoid function.
In Pix2Vox++/F, the output channels of transposed convolutional layers are numbered $128$, $64$, $32$, $8$ and $1$, respectively.
In Pix2Vox++/A, the output channel numbers of the five transposed convolutional layers are $512$, $128$, $32$, $8$ and $1$, respectively.
The decoder outputs a $32^3$ voxelized shape in the object's canonical view.

\noindent \textbf{High-resolution Reconstruction.}
To generate 3D volumes at $64^3$ resolution, there are six transposed convolutional layers in the decoders of Pix2Vox++/F and Pix2Vox++/A.
Each convolutional layer is with a batch normalization and a ReLU activation except for the last layer followed by a sigmoid function.
In Pix2Vox++/F, the output channels of the six transposed convolutional layers are numbered $128$, $64$, $32$, $16$, $8$ and $1$, respectively.
In Pix2Vox++/A, the output channel numbers of transposed convolutional layers are $512$, $128$, $32$, $16$, $8$ and $1$, respectively.
To generate 3D volumes at $128^3$ resolution, there are seven transposed convolutional layers in the decoders of Pix2Vox++/F and Pix2Vox++/A.
For Pix2Vox++/F, the output channels of the seven transposed convolutional layers are numbered $128$, $64$, $32$, $32$, $32$, $8$ and $1$, respectively.
For Pix2Vox++/A, the output channel numbers of the seven transposed convolutional layers are $512$, $128$, $32$, $32$, $32$, $8$ and $1$, respectively.

\subsubsection{Multi-scale Context-aware Fusion}

\begin{figure}[!t]
  \centering
  \resizebox{\linewidth}{!} {
    \includegraphics{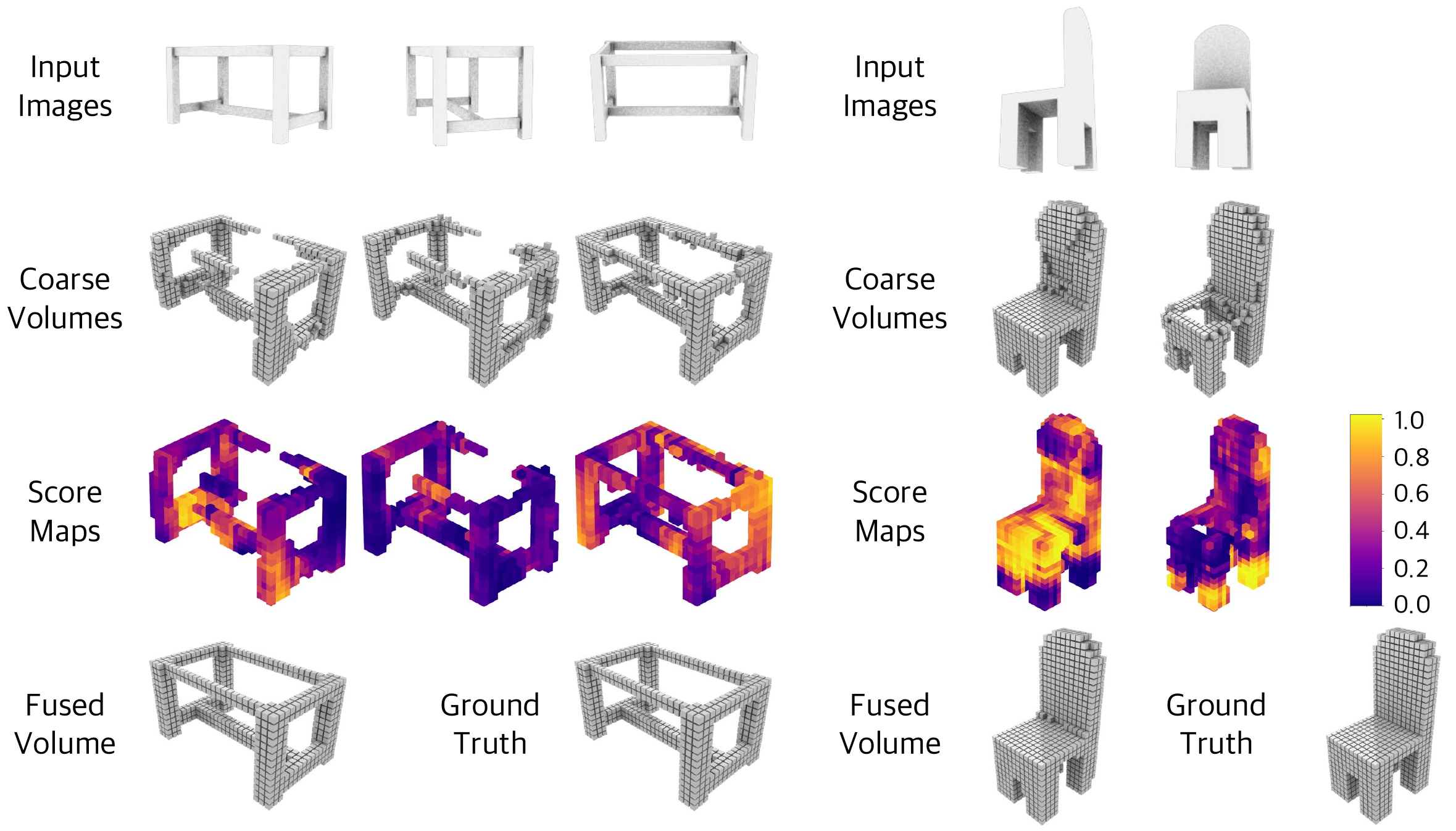}
  }
  \caption{Visualization of the score maps in the multi-scale context-aware fusion module.}
  \label{fig:context-aware-fusion-examples}
\end{figure}

Different parts of an object can be seen from different viewpoints.
The reconstruction qualities of visible parts are much higher than those of invisible parts.
Inspired by this observation, we propose a multi-scale context-aware fusion module to adaptively select high-quality reconstruction for each part from different coarse 3D volumes.
The selected reconstructions are fused to produce a 3D volume of the entire object.
As shown in Figure \ref{fig:context-aware-fusion-examples}, the multi-scale context-aware fusion generates higher scores for high-quality reconstructions, which can eliminate the effect of the missing of the wrongly recovered parts.

\begin{figure}[!t]
  \centering
  \resizebox{\linewidth}{!} {
    \includegraphics{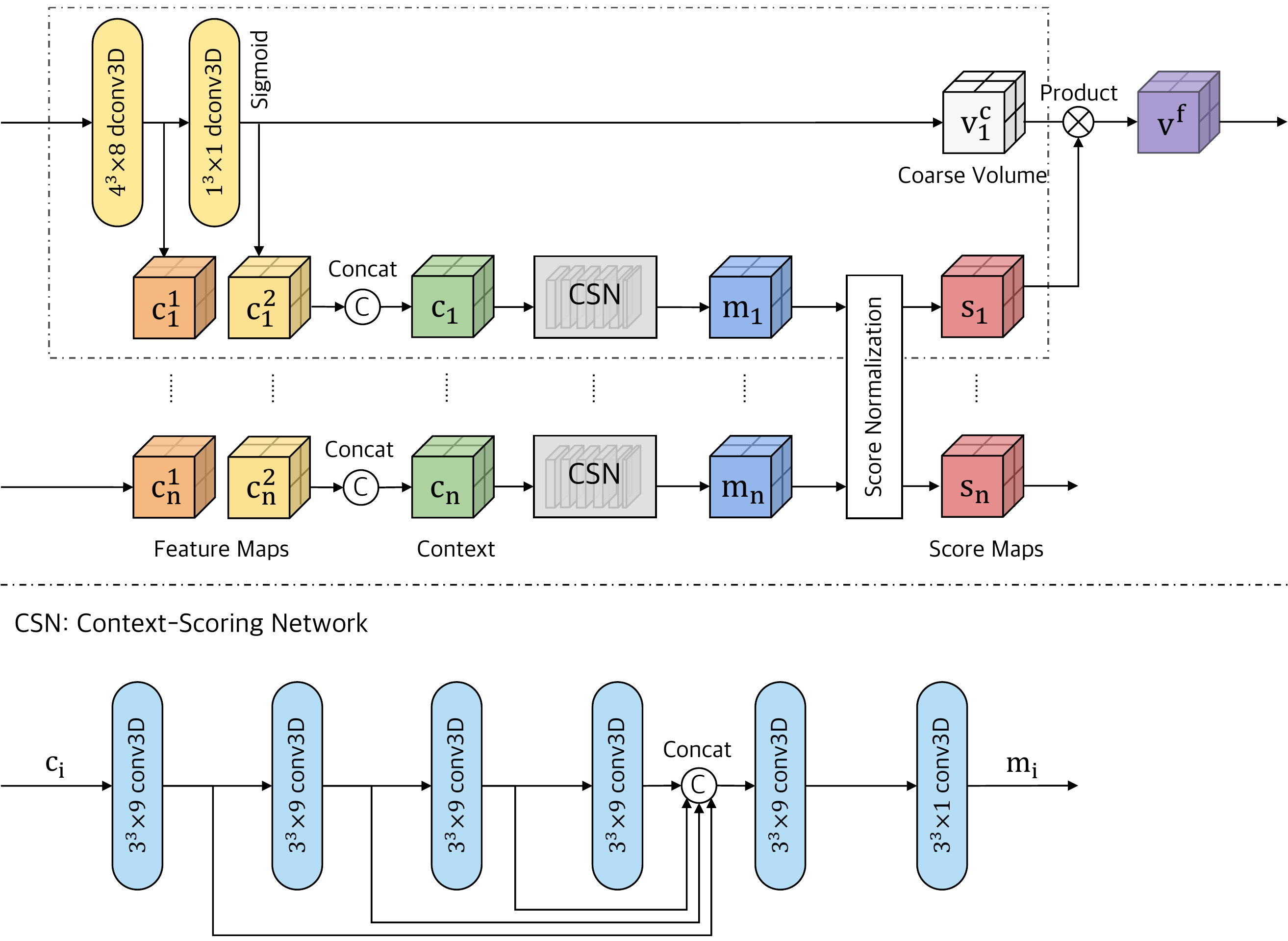}
  }
  \caption{An overview of the multi-scale context-aware fusion module. It aims to select high-quality reconstructions for each part to construct the final results. The objects in the bounding box describe the procedure score calculation for a coarse volume $v^c_1$. The other scores are calculated according to the same procedure. Note that the weights of the context scoring network are shared among different views.}
  \label{fig:context-aware-fusion-network-architecture}
\end{figure}

As shown in Figure \ref{fig:context-aware-fusion-network-architecture}, given coarse 3D volumes and the corresponding context, the multi-scale context-aware fusion module generates a score map for each coarse volume and then fuses them into one volume by weighted summation of all coarse volumes according to their score maps.
The spatial information of voxels is preserved in the multi-scale context-aware fusion module, and thus Pix2Vox++ can use multi-view information to recover the structure of an object better.

Deeper convolutional layers have larger receptive fields, which help to explore contextual information in 3D volumes.
However, the features in deeper convolutional layers may lose details of the object.
To address this problem, we concatenate multiple feature maps with different scales to preserve details in shallower convolutional layers, which is important for recovering details in high-resolution 3D volumes.
Specifically, the multi-scale context-aware fusion generates the context $c_r$ of the $r$-th coarse volume $v_r^c$ by concatenating the output of the last two layers in the decoder.
The context scoring network then generates a score $m_r$ for the context of the $r$-th coarse voxel.
The context scoring network consists of five sets of 3D convolutional layers, each with a kernel size of $3^3$ and padding of $1$, followed by a batch normalization and a leaky ReLU activation.
The numbers of output channels of the convolutional layers are $9$ except for the last layer whose number of output channels is $1$.
The feature maps generated by the first four convolutional layers are concatenated and forwarded to the fifth convolutional layer.
The learned score $m_r$ for context $c_r$ is normalized across all learned scores.
We choose softmax as the normalization function. 
Therefore, the score $s_r^{(i, j, k)}$ at the position $(i, j, k)$ for the $r$-th voxel can be calculated as 
\begin{equation}
  s_r^{(i, j, k)} = \frac{\exp\left(m_r^{(i, j, k)}\right)}{\sum_{p=1}^n \exp\left(m_p^{(i, j, k)}\right)}  
\end{equation}
where $n$ represents the number of views.
Finally, the fused voxel $v^f$ is produced by summing up the product of coarse voxels and the corresponding scores altogether.

\begin{equation}
  v^f = \sum_{r=1}^n s_r v_r^c
\end{equation}

\subsubsection{Refiner}

The refiner can be seen as a residual network, which aims to correct the wrongly recovered parts of a 3D volume.
It follows the concept of a 3D encoder-decoder with U-net connections \citep{DBLP:conf/miccai/RonnebergerFB15} that preserves the local structure in the fused volume.

\noindent \textbf{Low-resolution Reconstruction.}
To generate 3D volumes at $32^3$ resolution, the encoder has three 3D convolutional layers, each with a bank of $4^3$ filters with padding of $2$, followed by a batch normalization layer, a leaky ReLU activation and a max pooling layer with kernel size of $2^3$.
The output channels of convolutional layers are numbered $32$, $64$ and $128$, respectively.
The encoder is finally followed by two fully connected layers with dimensions of $2048$ and $8192$.
The decoder consists of three transposed convolutional layers, each with a bank of $4^3$ filters with padding of $2$ and stride of $1$.
Except for the last transposed convolutional layer which is followed by a sigmoid function, other layers are followed by a batch normalization layer, and a ReLU activation.
The numbers of output channels of the transposed convolutional layers are $64$, $32$ and $1$, respectively.

\noindent \textbf{High-resolution Reconstruction.}
To generate 3D volumes at $64^3$ resolution, we add a 3D convolutional layer before the first convolutional layer and a 3D transposed convolutional layer before the last layer.
Both layers are with output channels of $16$, kernel sizes of $4^3$, paddings of $2$, followed by batch normalization layers and leaky ReLU activations.
There is a max pooling layer with a kernel size of $2^3$ after the added 3D convolutional layer.
To generate 3D volumes at $128^3$ resolution, there are five 3D convolutional layers in the encoder, whose output channels are $8$, $16$, $32$, $64$ and $128$, respectively.
In the decoder, there are five 3D transposed convolutional layers with kernel sizes of $4^3$, paddings of $2$, and stride of $1$.
The output channels of the five layers are $64$, $32$, $16$, $8$ and $1$, respectively.

\subsection{Loss Function}

The loss function of the network is defined as the mean value of the voxel-wise binary cross entropies between the reconstructed object and the ground truth.
More formally, it can be defined as

\begin{equation}
  \ell = \frac{1}{N} \sum_{i=1}^N \left[ gt_i \log(p_i) + (1 - gt_i) \log(1 - p_i) \right]
  \label{eq:bce-loss}
\end{equation}
where $N$ denotes the number of voxels in the ground truth. $p_i$ and $gt_i$ represent the predicted occupancy and the corresponding ground truth.
The smaller the $\ell$ value is, the closer the prediction is to the ground truth.

\section{The Proposed Dataset: Things3D}
\label{sec:things3d}

There are several datasets available for 3D object reconstruction, including ShapeNet \citep{DBLP:conf/cvpr/WuSKYZTX15} and Pix3D \citep{DBLP:conf/cvpr/Sun0ZZZXTF18}.
ShapeNet is a large dataset for 3D models, but does not contain naturalistic background images.
Pix3D has real images, but it only contains 395 3D models and 10,069 images, which is not enough to train networks \citep{DBLP:conf/cvpr/TatarchenkoRRLK19}.

To generate a large-scale dataset for 3D object reconstruction with naturalistic backgrounds, \cite{DBLP:conf/iccv/SuQLG15} and \cite{DBLP:conf/cvpr/LinWRSKFL19} randomly warp and crop spherical images from the SUN database \citep{DBLP:conf/cvpr/XiaoHEOT10} and SUN360 database \citep{DBLP:conf/cvpr/XiaoEOT12} to sample background images, respectively.
Consequently, multi-view images are obtained by compositing foreground and background images together at the corresponding camera poses.
In contrast, we generate more realistic images from diverse virtual scenes with Blender, an open-source 3D creation suite, where we can easily control camera pose and location, as well as lighting conditions.

\begin{figure}[!t]
  \centering
  \resizebox{\linewidth}{!} {
    \includegraphics{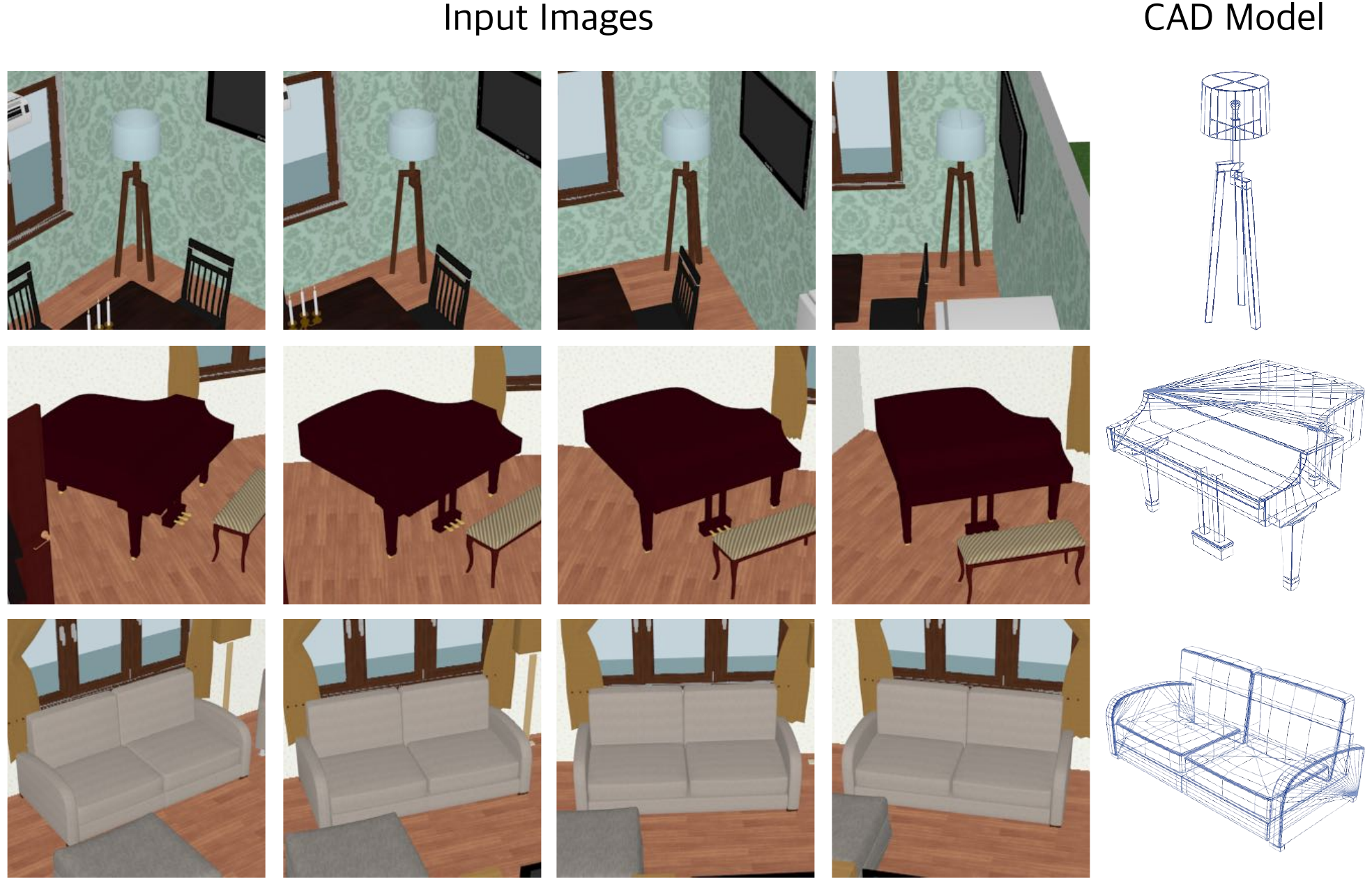}
  }
  \caption{Sample images and the corresponding CAD models in Things3D, where each 3D model is rendered from a diverse set of naturalistic scenes.}
  \label{fig:Things3D-samples}
\end{figure}

We present a large-scale naturalistic dataset for 3D object reconstruction, named Things3D, which contains 1.68M images of 280K objects (with 21K unique objects) collected from over 39K SUNCG \citep{DBLP:conf/cvpr/SongYZCSF17} indoor scenarios.
Sample images and corresponding CAD models are shown in Figure \ref{fig:Things3D-samples}.
To increase the diversity of 3D objects, we use 3D models in the ShapeNet dataset instead of the ones in the SUNCG dataset.
In particular, each 3D model in SUNCG scenes is replaced with one that is randomly selected from the same category in the ShapeNet dataset.
In addition, the replaced 3D model is of equal or smaller size than the original.
We randomly sample 24 viewing spheres for each object with yaw $\in [0, 360)$, pitch $= 30$, roll $= 0$ degrees.
The distance between the camera and the object is set to 10 unit length.
The camera has a focal length of 96 mm.
The power of light is uniformly sampled from $[500, 2000]$, and the specular of light is randomly sampled from $[0.75, 3]$.
Both the camera and the light track to the rendered object.
Specially, we ignore rendered images if more than 12.5\% of the object is occluded.
The images are of resolution $256 \times 256$.
In addition to rendered images, information about the canonical orientation and ground truth for objects is provided as well.
The data generation process lasts 32 days and runs on 15 servers with 4 Intel Xeon E5-2682 v4@2.50GHz CPUs and 256 GB RAM.
The dataset is available at \url{https://gateway.haozhexie.com/?fileName=Things3D}.

\section{Experiments}

\begin{table*}[!t]
  \caption{Comparison of single-view 3D object reconstruction on ShapeNet at $32^3$ resolution. We report the mean IoU per category. The best number for each category is highlighted in bold.}
  \resizebox{\linewidth}{!} {
    \begin{tabular}{lcccccccccc}
      \toprule
      Category     & 3D-R2N2    & OGN        & Matryoshka
                   & AtlasNet   & Pixel2Mesh & OccNet
                   & IM-Net     & AttSets    
                   & Pix2Vox++/F& Pix2Vox++/A \\
      \midrule
      airplane     & 0.513      & 0.587      & 0.647
                   & 0.493      & 0.508      & 0.532
                   & \bf{0.702} & 0.594
                   & 0.607      & 0.674 \\
      bench        & 0.421      & 0.481      & 0.577
                   & 0.431      & 0.379      & 0.597
                   & 0.564      & 0.552
                   & 0.544      & \bf{0.608} \\
      cabinet      & 0.716      & 0.729      & 0.776
                   & 0.257      & 0.732      & 0.674
                   & 0.680      & 0.783
                   & 0.782      & \bf{0.799} \\
      car          & 0.798      & 0.828      & 0.850
                   & 0.282      & 0.670      & 0.671
                   & 0.756      & 0.844
                   & 0.841      & \bf{0.858} \\
      chair        & 0.466      & 0.483      & 0.547
                   & 0.328      & 0.484      & 0.583
                   & \bf{0.644} & 0.559
                   & 0.548      & 0.581 \\
      display      & 0.468      & 0.502      & 0.532
                   & 0.457      & 0.582      & \bf{0.651}
                   & 0.585      & 0.565
                   & 0.529      & 0.548 \\
      lamp         & 0.381      & 0.398      & 0.408
                   & 0.261      & 0.399      & \bf{0.474}
                   & 0.433      & 0.445
                   & 0.448      & 0.457 \\
      speaker      & 0.662      & 0.637      & 0.701
                   & 0.296      & 0.672      & 0.655
                   & 0.683      & 0.721
                   & 0.721      & \bf{0.721} \\
      rifle        & 0.544      & 0.593      & 0.616
                   & 0.573      & 0.468      & 0.656
                   & \bf{0.723} & 0.601
                   & 0.594      & 0.617 \\
      sofa         & 0.628      & 0.646      & 0.681
                   & 0.354      & 0.622      & 0.669
                   & 0.694      & 0.703
                   & 0.696      & \bf{0.725} \\
      table        & 0.513      & 0.536      & 0.573
                   & 0.301      & 0.536      & \bf{0.659}
                   & 0.621      & 0.590
                   & 0.609      & 0.620 \\
      telephone    & 0.661      & 0.702      & 0.756
                   & 0.543      & 0.762      & 0.794
                   & 0.762      & 0.743
                   & 0.782      & \bf{0.809} \\
      watercraft   & 0.513      & 0.632      & 0.591
                   & 0.355      & 0.471      & 0.579
                   & \bf{0.607} & 0.601
                   & 0.583      & 0.603 \\
      \midrule
      Overall      & 0.560      & 0.596      & 0.635
                   & 0.352      & 0.552      & 0.626
                   & 0.659      & 0.642
                   & 0.645      & \bf{0.670} \\
      \bottomrule
    \end{tabular}
  }
  \label{tab:shapenet-voxel-reconstruction-iou}
\end{table*}

\begin{table*}[!t]
  \caption{Comparison of single-view 3D object reconstruction on ShapeNet. We report the mean F-Score@1\% per category. For voxel reconstruction methods, the points are sampled from triangular meshes generated by the marching cube algorithm. The best number for each category is highlighted in bold.}
  \resizebox{\linewidth}{!} {
    \begin{tabular}{lcccccccccc}
      \toprule
      Category     & 3D-R2N2    & OGN        & Matryoshka
                   & AtlasNet   & Pixel2Mesh & OccNet
                   & IM-Net     & AttSets    
                   & Pix2Vox++/F& Pix2Vox++/A \\
      \midrule
      airplane     & 0.412      & 0.487      & 0.446
                   & 0.415      & 0.376      & 0.494
                   & \bf{0.598} & 0.489
                   & 0.493      & 0.583 \\
      bench        & 0.345      & 0.364      & 0.424
                   & 0.439      & 0.313      & 0.318
                   & 0.361      & 0.406
                   & 0.399      & \bf{0.478} \\
      cabinet      & 0.327      & 0.316      & 0.381
                   & 0.350      & \bf{0.450} & 0.449
                   & 0.345      & 0.367
                   & 0.363      & 0.408 \\
      car          & 0.481      & 0.514      & 0.481
                   & 0.319      & 0.486      & 0.315
                   & 0.304      & 0.497
                   & 0.523      & \bf{0.564} \\
      chair        & 0.238      & 0.226      & 0.302
                   & 0.406      & 0.386      & 0.365
                   & \bf{0.442} & 0.334
                   & 0.262      & 0.309 \\
      display      & 0.227      & 0.215      & 0.400
                   & 0.451      & 0.319      & \bf{0.468}
                   & 0.466      & 0.310
                   & 0.253      & 0.296 \\
      lamp         & 0.267      & 0.249      & 0.276
                   & 0.217      & 0.219      & 0.361
                   & \bf{0.371} & 0.315
                   & 0.287      & 0.315 \\
      speaker      & 0.231      & 0.225      & 0.279
                   & 0.199      & 0.190      & \bf{0.249}
                   & 0.200      & 0.211
                   & 0.256      & 0.152 \\
      rifle        & 0.521      & 0.541      & 0.514
                   & 0.405      & 0.340      & 0.219
                   & 0.407      & 0.524
                   & 0.553      & \bf{0.574} \\
      sofa         & 0.274      & 0.290      & 0.326
                   & 0.337      & 0.343      & 0.324
                   & 0.354      & 0.334
                   & 0.320      & \bf{0.377} \\
      table        & 0.340      & 0.352      & 0.374
                   & 0.373      & 0.502      & \bf{0.549}
                   & 0.461      & 0.419
                   & 0.385      & 0.406 \\
      telephone    & 0.504      & 0.528      & 0.598
                   & 0.545      & 0.485      & 0.273
                   & 0.423      & 0.469
                   & 0.588      & \bf{0.633} \\
      watercraft   & 0.305      & 0.328      & 0.360
                   & 0.296      & 0.266      & 0.347
                   & 0.369      & 0.315
                   & 0.346      & \bf{0.390} \\
      \midrule
      Overall      & 0.351      & 0.368      & 0.391
                   & 0.362      & 0.398      & 0.393
                   & 0.405      & 0.395
                   & 0.394      & \bf{0.436} \\
      \bottomrule
    \end{tabular}
  }
  \label{tab:shapenet-voxel-reconstruction-fscore}
\end{table*}

\begin{figure*}
  \centering
  \resizebox{\linewidth}{!} {
    \includegraphics{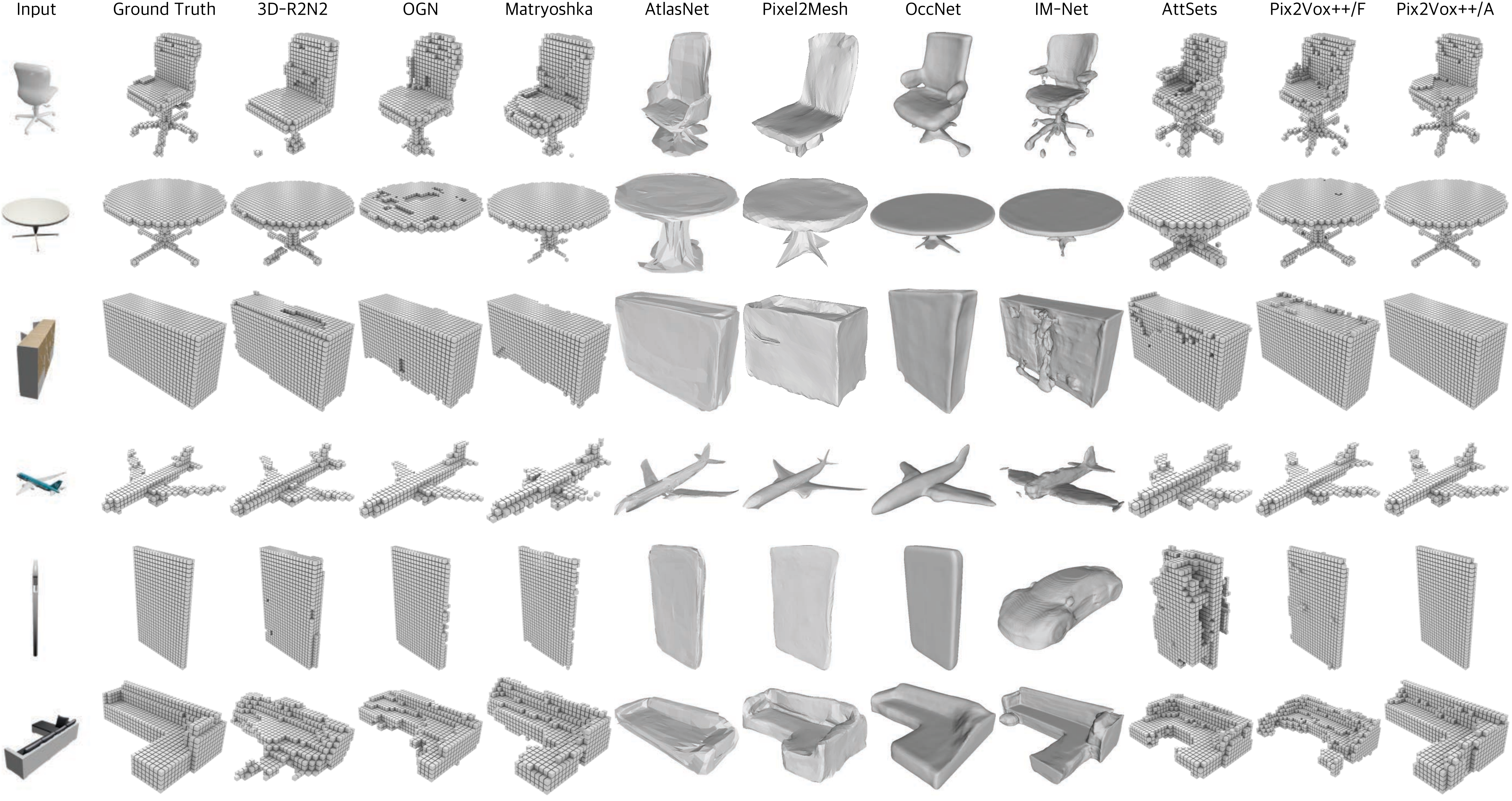}
  }
  \caption{Example of single-view 3D object reconstruction on ShapeNet. For voxel reconstruction methods, the output 3D volumes are at $32^3$ resolution.}
  \label{fig:shapenet-single-view-reconstruction}
  \vspace{1 mm}
\end{figure*}

In this section, we present extensive experimental evaluations of Pix2Vox++ on the ShapeNet \citep{DBLP:conf/cvpr/WuSKYZTX15}, Pix3D \citep{DBLP:conf/cvpr/Sun0ZZZXTF18}, and Things3D datasets.
We first describe the datasets and evaluation protocols.
Next, we demonstrate the implementation details of the proposed methods.
Finally, we report experimental evaluations of the proposed methods against state-of-the-art methods.

\subsection{Datasets}

\noindent \textbf{ShapeNet.}
The ShapeNet dataset \citep{DBLP:conf/cvpr/WuSKYZTX15} is a collection of 3D CAD models organized according to the WordNet taxonomy.
We use a subset of the ShapeNet dataset consisting of 44K models and 13 major categories following \cite{DBLP:conf/eccv/ChoyXGCS16}.
More specifically, we use renderings provided by 3D-R2N2 which contains 24 random views of size $137 \times 137$ for each 3D model.
We also apply a uniform colored background to the image during training and testing.

\noindent \textbf{Pix3D.}
The Pix3D dataset \citep{DBLP:conf/cvpr/Sun0ZZZXTF18} provides perfectly aligned real-world images and CAD models.
The dataset contains 395 3D models of nine object classes.
Each model is associated with a set of real images, capturing the exact object in diverse environments.
Following \cite{DBLP:conf/cvpr/Sun0ZZZXTF18}, we use 2,894 untruncated and unoccluded images from the chair category for testing in the following experiments.

\noindent \textbf{Things3D.}
The Things3D dataset proposed in this paper contains 1.68M images of 280K objects collected from over 39K SUNCG \citep{DBLP:conf/cvpr/SongYZCSF17} indoor scenarios.
See Section \ref{sec:things3d} for more details.

\subsection{Metrics}

To evaluate the reconstruction quality of the proposed methods, we binarize probabilities at a fixed threshold of 0.3 and use intersection over union (IoU) as a similarity measure between prediction and ground truth.
More formally,

\begin{equation}
  {\rm IoU} = \frac{\sum_{i, j, k} {\rm I}(\hat{p}_{(i, j, k)} > t) {\rm I}(\hat{p}_{(i, j, k)})}{\sum_{i, j, k} {\rm I}\left[ {\rm I}(\hat{p}_{(i, j, k)} > t) + {\rm I}(p_{(i, j, k)}) \right]}
\end{equation}
where $\hat{p}_{(i, j, k)}$ and $p_{(i, j, k)}$ represent the predicted occupancy probability and ground truth at $(i, j, k)$, respectively.
${\rm I}(\cdot)$ is an indicator function and $t$ denotes a voxelization threshold.
Higher IoU values indicate better reconstruction results.

Following \cite{DBLP:conf/cvpr/TatarchenkoRRLK19}, we also take F-Score as an extra metric to evaluate the performance of 3D reconstruction results, which can be defined as

\begin{equation}
  \textnormal{F-Score}(d) = \frac{2P(d)R(d)}{P(d) + R(d)}
\end{equation}
where $P(d)$ and $R(d)$ denote the precision and recall for a distance threshold $d$, respectively.
They can be computed as 

\begin{equation}
  P(d) = \frac{1}{n_{\mathcal{R}}} \sum_{r \in \mathcal{R}} \left[\min_{g \in \mathcal{G}} ||g - r|| < d \right]
\end{equation}

\begin{equation}
  R(d) = \frac{1}{n_{\mathcal{G}}} \sum_{g \in \mathcal{G}} \left[\min_{r \in \mathcal{R}} ||g - r|| < d \right]
\end{equation}
where $\mathcal{R}$ and $\mathcal{G}$ denote the reconstructed and ground truth point clouds, respectively.
$n_\mathcal{R}$ and $n_\mathcal{G}$ are the numbers of points in $\mathcal{R}$ and $\mathcal{G}$, respectively. 
For voxel reconstruction methods, we first apply the marching cubes algorithm \citep{DBLP:conf/siggraph/LorensenC87} to generate the object surface.
We then sample 8,192 points from the object surface to compute F-Score between prediction and ground truth.
For mesh and signed distance field reconstruction methods, we also sample 8,192 points from the object surface to compute F-Score.
Higher F-Score values indicate better reconstructions.

\subsection{Implementation Details}

We train the proposed methods with batch size $64$ using $224 \times 224$ RGB images as input.
The output voxelized reconstruction is $32^3$ in size.
We implement our networks in PyTorch \citep{DBLP:conf/nips/AdamSSGEZZALA17} and train both Pix2Vox++/F and Pix2Vox++/A using an Adam optimizer \citep{DBLP:conf/iclr/KingmaB15} with a $\beta_1$ of $0.9$ and a $\beta_2$ of $0.999$.
The initial learning rate is set to $0.001$ and decayed by 2 after 150 epochs.
First, we train networks except for the multi-scale context-aware fusion using single-view images for 250 epochs.
We then train the networks using multi-view images for 100 epochs.

\subsection{Evaluation on the ShapeNet Dataset}

\begin{table*}[!t]
  \caption{Comparison of multi-view 3D object reconstruction on ShapeNet at $32^3$ resolution. We report the mean IoU and F-Score@1\% for all categories.}
  \centering
  \begin{tabularx}{\linewidth}{lYYYYYYccc}
    \toprule
    Methods (IoU)
                & 1 view         & 2 views        & 3 views
                & 4 views        & 5 views        & 8 views
                & 12 views       & 16 views       & 20 views \\
    \midrule
    \multicolumn{10}{l}{\bf{Metric: IoU}} \\
    \midrule
    3D-R2N2     & 0.560          & 0.603          & 0.617
                & 0.625          & 0.634          & 0.635
                & 0.636          & 0.636          & 0.636 \\
    AttSets     & 0.642          & 0.662          & 0.670
                & 0.675          & 0.677          & 0.685
                & 0.688          & 0.692          & 0.693 \\
    Pix2Vox++/F & 0.645          & 0.669          & 0.678
                & 0.682          & 0.685          & 0.690
                & 0.692          & 0.693          & 0.694\\
    Pix2Vox++/A & \bf{0.670}     & \bf{0.695}     & \bf{0.704}
                & \bf{0.708}     & \bf{0.711}     & \bf{0.715}
                & \bf{0.717}     & \bf{0.718}     & \bf{0.719} \\
    \midrule
    \midrule
    \multicolumn{10}{l}{\bf{Metric: F-Score@1\%}} \\
    \midrule
    3D-R2N2     & 0.351          & 0.368          & 0.372
                & 0.378          & 0.382          & 0.383
                & 0.382          & 0.382          & 0.383 \\
    AttSets     & 0.395          & 0.418          & 0.426
                & 0.430          & 0.432          & 0.444
                & 0.445          & 0.447          & 0.448 \\
    Pix2Vox++/F & 0.394          & 0.422          & 0.432
                & 0.437          & 0.440          & 0.446
                & 0.449          & 0.450          & 0.451 \\
    Pix2Vox++/A & \bf{0.436}     & \bf{0.452}     & \bf{0.455}
                & \bf{0.457}     & \bf{0.458}     & \bf{0.459}
                & \bf{0.460}     & \bf{0.461}     & \bf{0.462} \\
    \bottomrule
  \end{tabularx}
  \label{tab:shapenet-multi-view-reconstruction}
  \vspace{2 mm}
\end{table*}

\subsubsection{Single-view 3D Object Reconstruction}

To evaluate the performance of the proposed methods in handling clean background images, we compare our methods against several state-of-the-art methods, including 3D-R2N2 \citep{DBLP:conf/eccv/ChoyXGCS16}, OGN \citep{DBLP:conf/iccv/TatarchenkoDB17}, Matryoshaka \citep{DBLP:conf/cvpr/Richter018}, AtlasNet \citep{DBLP:conf/cvpr/GroueixFKRA18}, Pixel2Mesh \citep{DBLP:conf/eccv/WangZLFLJ18}, OccNet \citep{DBLP:conf/cvpr/MeschederONNG19}, IM-Net \citep{DBLP:conf/cvpr/ChenZ19}, and AttSets \citep{DBLP:journals/ijcv/YangSAN19}.
Tables \ref{tab:shapenet-voxel-reconstruction-iou} and \ref{tab:shapenet-voxel-reconstruction-fscore} show the IoU and F-Score@1\% of all methods on the ShapeNet test set.
Both Pix2Vox++/F and Pix2Vox++/A outperform all competitive methods in terms of both IoU and F-Score@1\%.
Figure \ref{fig:shapenet-single-view-reconstruction} provides qualitative results for single-view reconstruction results showing that Pix2Vox++/F and Pix2Vox++/A generate more visually compelling 3D shapes than other methods.

\subsubsection{Multi-view 3D Object Reconstruction}

\begin{figure}
  \centering
  \resizebox{\linewidth}{!} {
    \includegraphics{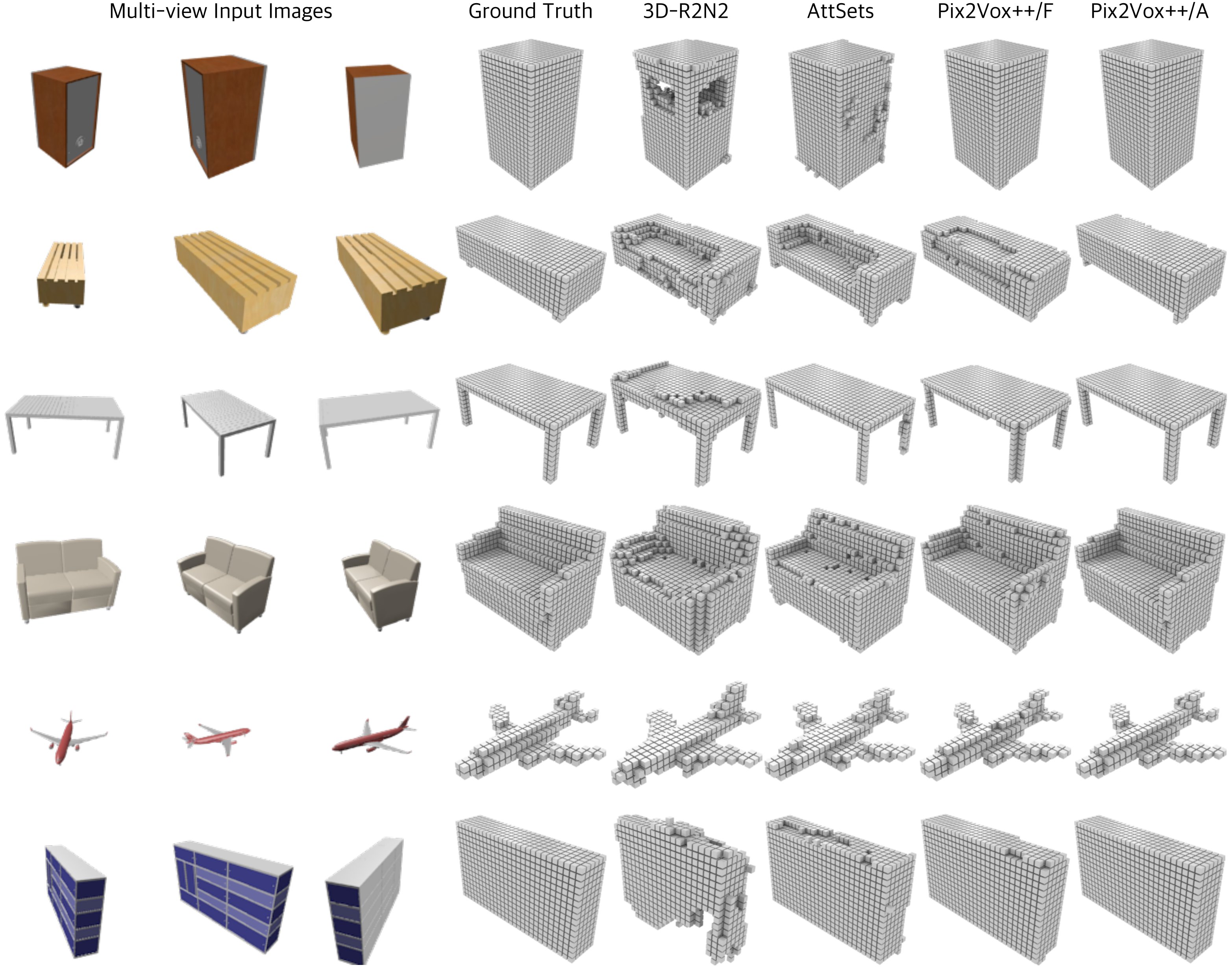}
  }
  \caption{Example of multi-view 3D object reconstruction on ShapeNet at $32^3$ resolution.}
  \label{fig:shapenet-multi-view-reconstruction}
\end{figure}

To evaluate the performance of reconstructing 3D objects from multi-view images, we compare the proposed methods with 3D-R2N2 \citep{DBLP:conf/eccv/ChoyXGCS16} and AttSets \citep{DBLP:journals/ijcv/YangSAN19}.
As shown in Table \ref{tab:shapenet-multi-view-reconstruction}, the proposed Pix2Vox++/F and Pix2Vox++/A consistently outperform 3D-R2N2 and AttSets in all numbers of views.
Figure \ref{fig:shapenet-multi-view-reconstruction} shows several examples reconstructed from three input images.
Both Pix2Vox++/F and Pix2Vox++/A are able to recover better details than other methods.
Pix2Vox++/A performs better in 3D object reconstruction by comparing with Pix2Vox++/F. 

To provide a detailed analysis of the multi-scale context-aware fusion, we visualize score maps of corresponding coarse volumes when reconstructing the 3D shape of a table and a chair, as shown in Figure \ref{fig:context-aware-fusion-examples}.
The chair seat on the right is of low quality, and the score of the corresponding part is lower than that in the other coarse volumes.
The fused 3D volume is obtained by combining selected high-quality reconstruction parts, where bad reconstructions can be effectively eliminated by our scoring scheme.

\subsubsection{Higher-Resolution 3D Object Reconstruction}

\begin{table*}[!t]
  \caption{Comparison of single-view and multi-view 3D object reconstruction on ShapeNet-Cars at $64^3$ and $128^3$ resolutions. We report the mean IoU and F-Score@1\% of all models.}
  \centering
  \resizebox{\linewidth}{!} {
    \begin{tabularx}{\linewidth}{lYYYY|YYYY}
        \toprule
        \multirow{2}{*}{Methods}
                   & \multicolumn{4}{c|}{IoU}
                   & \multicolumn{4}{c}{F-Score@1\%} \\
        \cline{2-9}
                   & 1 view    & 2 views   & 4 views   & 8 views
                   & 1 view    & 2 views   & 4 views   & 8 views \\
        \midrule
        \bf{Resolution: $64^3$} \\
        \midrule
        OGN        & 0.771     & N/A       & N/A       & N/A
                   & 0.361     & N/A       & N/A       & N/A \\
        Matryoshka & 0.784     & N/A       & N/A       & N/A
                   & 0.380     & N/A       & N/A       & N/A \\
        Pix2Vox++/F& 0.793     & 0.807     & 0.811     & 0.815
                   & 0.401     & 0.429     & 0.439     & 0.453 \\
        Pix2Vox++/A& \bf{0.803}& \bf{0.813}& \bf{0.815}& \bf{0.819} 
                   & \bf{0.418}& \bf{0.448}& \bf{0.450}& \bf{0.457}\\
        \midrule 
        \midrule
        \bf{Resolution: $128^3$}  \\
        \midrule
        OGN        & 0.782     & N/A       & N/A       & N/A
                   & 0.390     & N/A       & N/A       & N/A \\
        Matryoshka & 0.794     & N/A       & N/A       & N/A
                   & 0.426     & N/A       & N/A       & N/A \\
        Pix2Vox++/F& 0.817     & 0.832     & 0.838     & 0.840 
                   & 0.459     & 0.502     & 0.520     & 0.528 \\
        Pix2Vox++/A& \bf{0.826}& \bf{0.837}& \bf{0.841}& \bf{0.843} 
                   & \bf{0.475}& \bf{0.509}& \bf{0.521}& \bf{0.539}\\
        \bottomrule
    \end{tabularx}
  }
  \label{tab:shapenet-high-resolution-reconstruction}
\end{table*}

\begin{figure}[!t]
  \centering
  \resizebox{\linewidth}{!} {
    \includegraphics{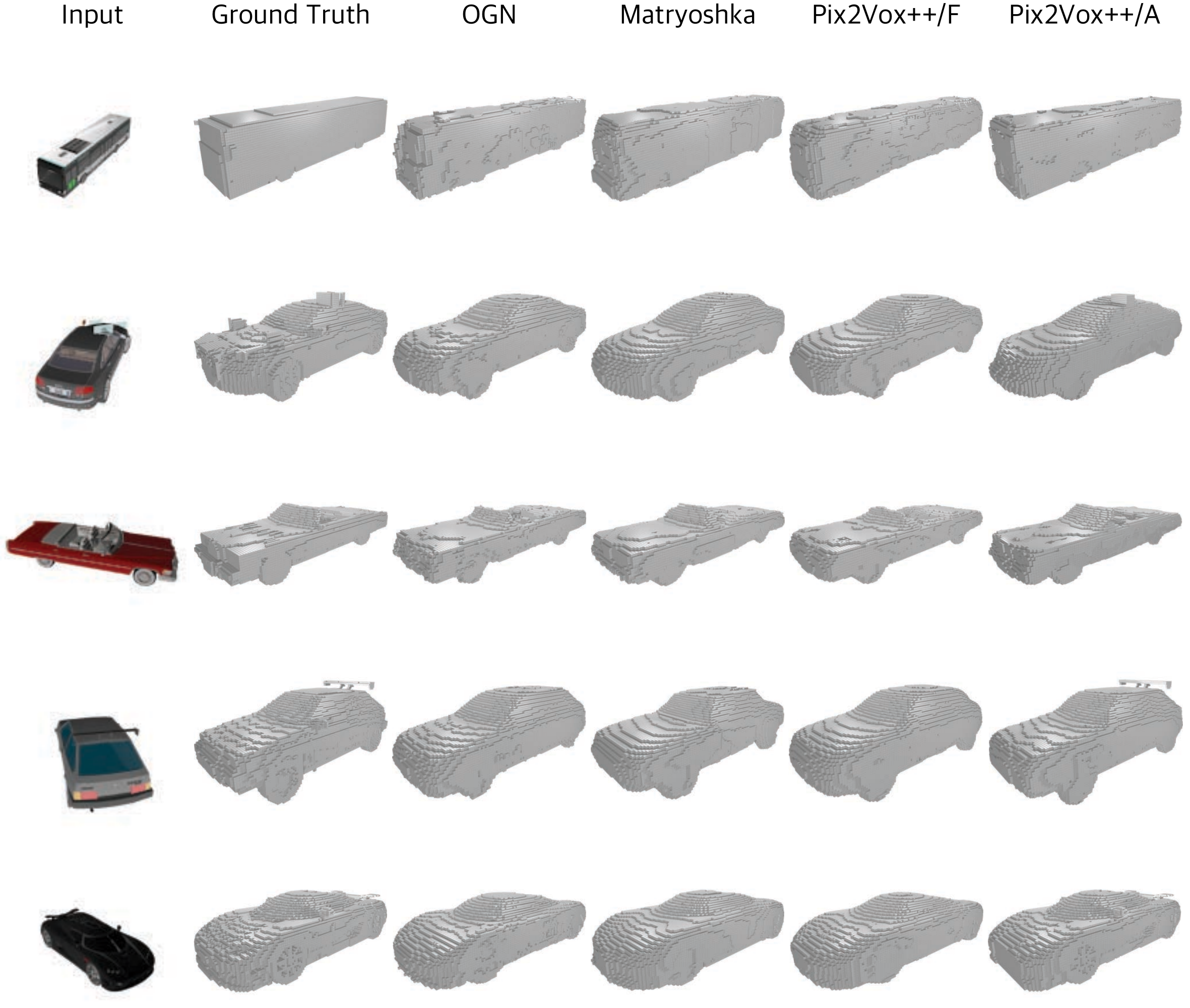}
  }
  \caption{Example of single-view 3D object reconstruction on ShapeNet-Cars at $128^3$ resolution.}
  \label{fig:shapenet-hd-reconstruction}
\end{figure}

Low-resolution 3D volumes are naturally limited to the low level of details they can represent.
To evaluate the performance of Pix2Vox++ in high-resolution 3D reconstruction, we compare it to Matryoshka Networks \citep{DBLP:conf/cvpr/Richter018} and OGN \citep{DBLP:conf/iccv/TatarchenkoDB17}.
We follow the experimental setup of OGN and predict 3D volumes of ShapeNet-Cars at $64^3$ and $128^3$ resolutions given a single RGB image.
We then upsample the predicted 3D volumes to $256^3$ resolution and compute IoU with ground truth shapes.
To make a fair comparison, we use the same dataset split and ground truth shapes generated by OGN.
We report quantitative results in Table \ref{tab:shapenet-high-resolution-reconstruction} and qualitative results at $128^3$ resolution in Figure \ref{fig:shapenet-hd-reconstruction}.
Experimental results on single-view reconstruction show that both Pix2Vox++/F and Pix2Vox++/A outperform Matryoshka Networks and OGN at $64^3$ resolution.
At $128^3$ resolution, Pix2Vox++/A outforms Matryoshka Networks and OGN.
Pix2Vox++/F is comparable to both competitive methods.
As shown in Figure \ref{fig:shapenet-hd-reconstruction}, our methods recover better details than compared methods. 
Furthermore, we provide results for high-resolution reconstruction from multi-view images.
Experimental results show that Pix2Vox++/A outperforms Pix2Vox++/F in all numbers of views.

\subsection{Evaluation on the Pix3D Dataset}

To evaluate the performance of the proposed methods on real-world images, we evaluate our methods for single-view reconstruction on the Pix3D dataset.
We train Pix2Vox++/F and Pix2Vox++/A on ShapeNet-Chairs and Things3D-Chairs and test both networks on the chair category of the Pix3D dataset.
As shown in Table \ref{tab:pix3d-reconstruction}, our networks trained on Things3D-Chairs have better results than those trained on ShapeNet-Chairs.

Following Pix3D \citep{DBLP:conf/cvpr/Sun0ZZZXTF18}, we use Render for CNN \citep{DBLP:conf/iccv/SuQLG15} to generate 60 images for each chair in the ShapeNet dataset by adding random backgrounds sampled from the SUN database \citep{DBLP:conf/cvpr/XiaoHEOT10}, {\it i.e.} ShapeNet-Chairs-RfC.
We also render a new dataset, called Things-3D-Chairs-RfC, by rendering chairs in the naturalist scenes where the camera poses are sampled from a distribution estimated on the Pix3D dataset.
Table \ref{tab:pix3d-reconstruction} illustrates that our networks trained on Things3D-Chairs-RfC have better results in reconstructing 3D objects in Pix3D than those trained on ShapeNet-Chairs-RfC.
The two above experiments show that the networks trained on datasets generated by rendering naturalist scenes archive better results than those trained on datasets with random backgrounds.
Furthermore, the proposed Pix2Vox++/A trained on Things3D-Chairs-RfC archives the best results on the Pix3D dataset, suggesting that the Thing3D dataset helps the networks generalize better to real-world datasets.

\begin{table}[!t]
  \caption{Comparison of single-view 3D object reconstruction on Pix3D at $32^3$ resolution. We report the mean IoU and F-Score@1\% of the chair category. The best number is highlighted in bold.}
  \centering
  \begin{tabularx}{\linewidth}{lYYYY}
    \toprule
  	Method       & IoU        & F-Score@1\% \\
  	\midrule
  	\multicolumn{3}{l}{\textbf{Training on} ShapeNet-Chairs} \\
  	\midrule
  	Pix2Vox++/F  & 0.179      & 0.012 \\
  	Pix2Vox++/A  & 0.204      & 0.018 \\
  	\midrule
  	\multicolumn{3}{l}{\textbf{Training on} Things3D-Chairs} \\
  	\midrule
  	Pix2Vox++/F  & 0.256      & 0.028 \\
  	Pix2Vox++/A  & 0.269      & 0.036 \\
  	\midrule
  	\midrule
  	\multicolumn{3}{l}{\textbf{Training on} ShapeNet-Chairs-RfC} \\
  	\midrule
  	Pix3D        & 0.282      & 0.041 \\
  	Pix2Vox++/F  & 0.276      & 0.042 \\
  	Pix2Vox++/A  & 0.292      & 0.068 \\
  	\midrule
  	\multicolumn{3}{l}{\textbf{Training on} Things3D-Chairs-RfC} \\
  	\midrule
  	Pix2Vox++/F  & 0.297      & 0.072 \\
  	Pix2Vox++/A  & \bf{0.324} & \bf{0.084} \\
  	\bottomrule
  \end{tabularx}
  \label{tab:pix3d-reconstruction}
\end{table}

\begin{figure}[!t]
  \centering
  \resizebox{\linewidth}{!} {
    \includegraphics{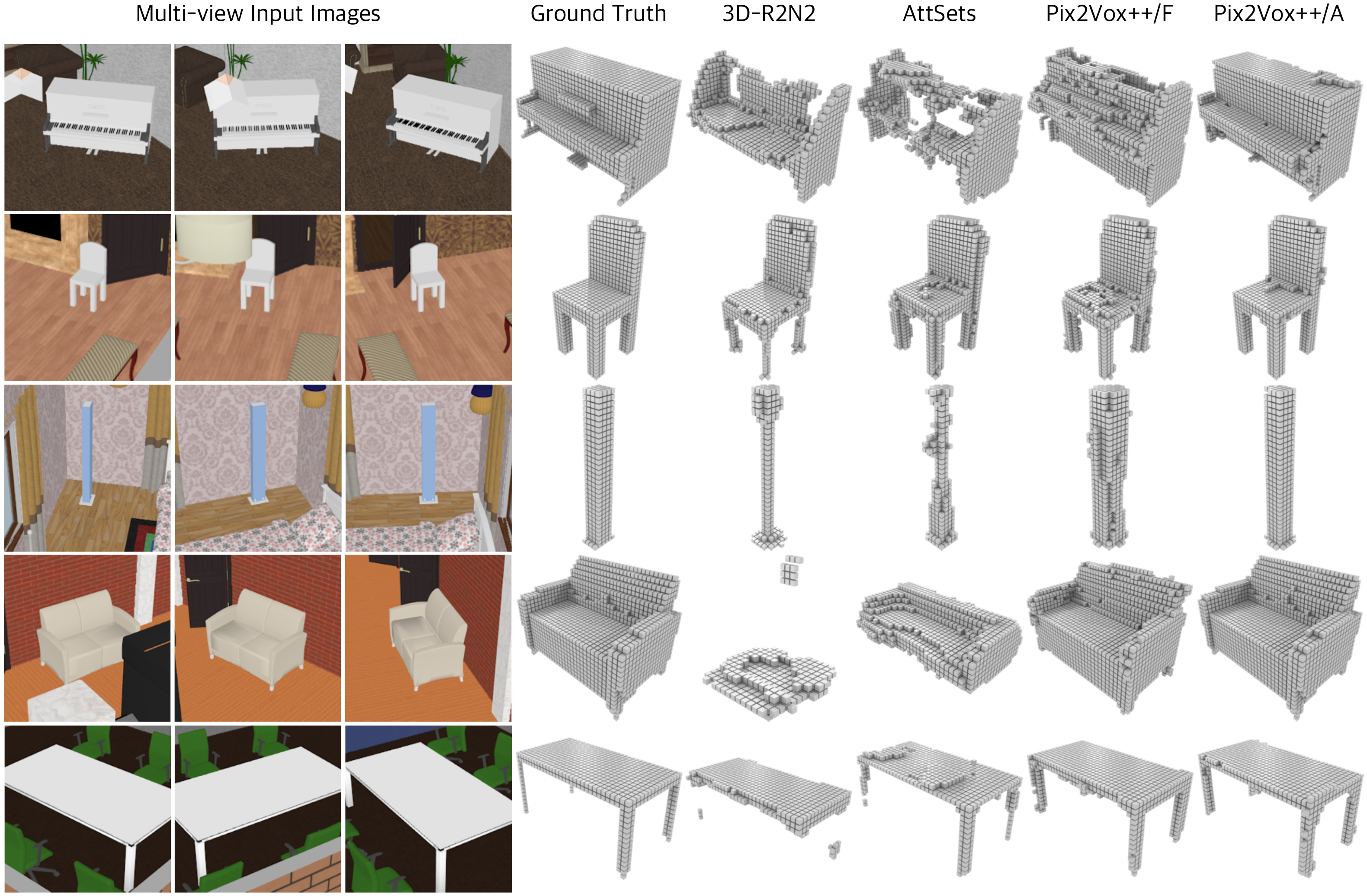}
  }
  \caption{Example of multi-view 3D object reconstruction on Things3D at $32^3$ resolution.}
  \label{fig:things3d-multi-view-reconstruction}
\end{figure}

\begin{figure}[!t]
  \centering
  \resizebox{\linewidth}{!} {
    \includegraphics{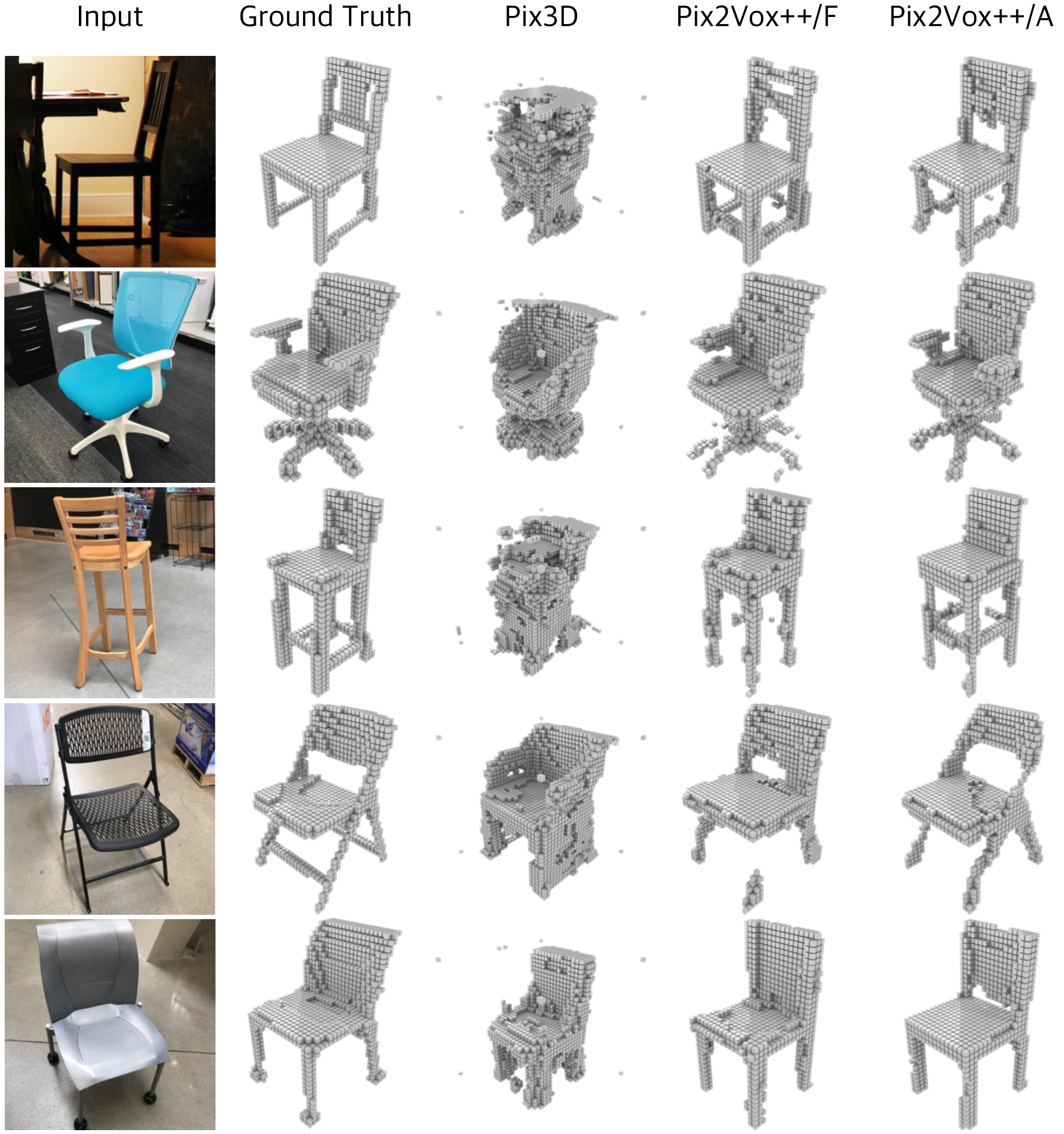}
  }
  \caption{Example of single-view 3D object reconstruction on Pix3D at $32^3$ resolution trained on ShapeNet-Chairs-RfC.}
  \label{fig:pix3d-reconstruction}
\end{figure}

\subsection{Evaluation on the Things3D Dataset}

\begin{figure*}
  \centering
  \resizebox{\linewidth}{!} {
    \includegraphics{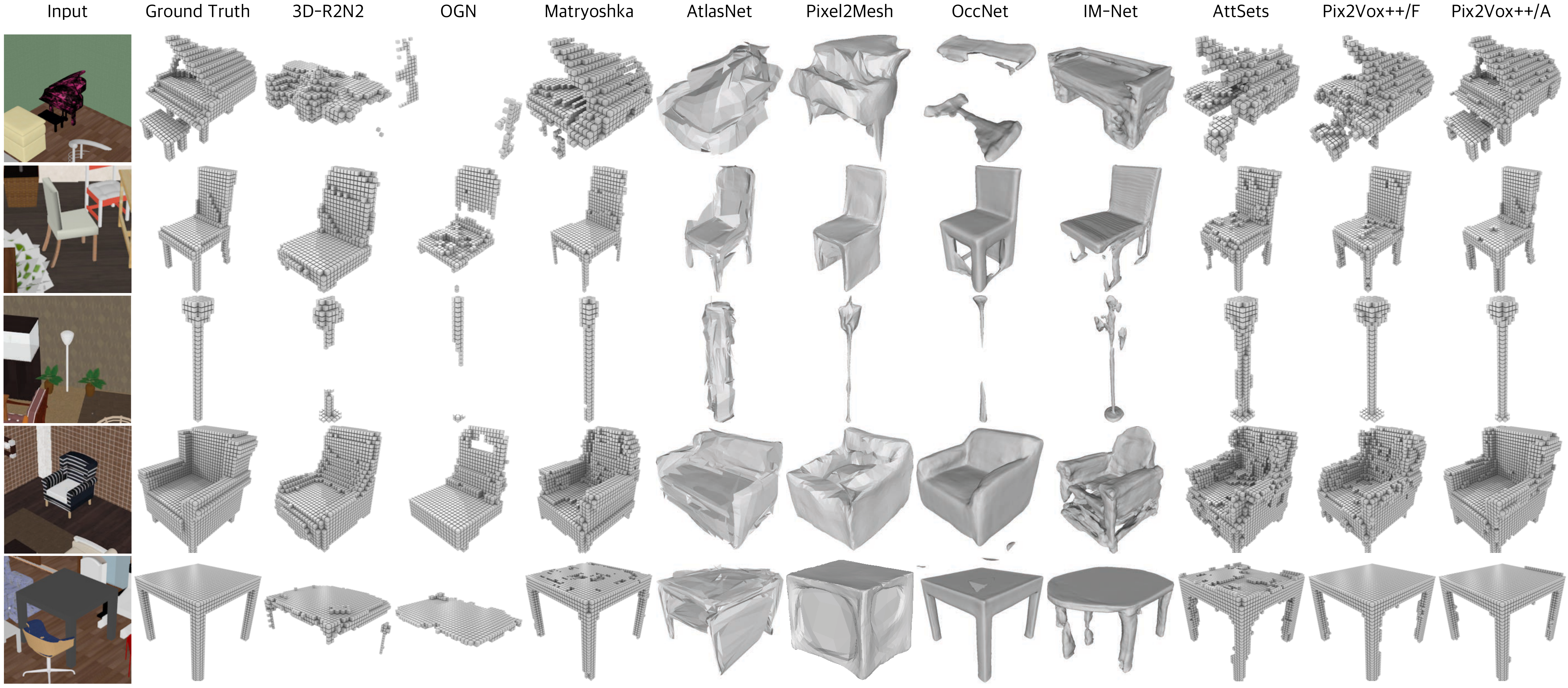}
  }
  \caption{Example of single-view 3D object reconstruction on Things3D. For voxel reconstruction methods, the output 3D volumes are at $32^3$ resolution.}
  \label{fig:things3d-single-view-reconstruction}
\end{figure*}

\begin{table*}[!t]
  \caption{Comparison of single-view 3D object reconstruction on Things3D at $32^3$ resolution. We report the mean IoU per category. The best number for each category is highlighted in bold.}
  \resizebox{\linewidth}{!} {
    \begin{tabular}{lcccccccccc}
      \toprule
      Category     & 3D-R2N2     & OGN        & Matryoshka
                   & AtlasNet    & Pixel2Mesh & OccNet
                   & IM-Net      & AttSets    
                   & Pix2Vox++/F & Pix2Vox++/A \\
      \midrule
      chair        & 0.327       & 0.212      & 0.399
                   & 0.251       & 0.372      & 0.432
                   & \bf{0.462}  & 0.403
                   & 0.435       & 0.442 \\
      display      & 0.240       & 0.153      & 0.332
                   & 0.259       & 0.311      & 0.328
                   & 0.324       & 0.301
                   & 0.324       & \bf{0.349} \\
      lamp         & 0.257       & 0.189      & 0.323
                   & 0.196       & 0.306      & 0.361
                   & 0.328       & 0.334
                   & 0.350       & \bf{0.362} \\
      piano        & 0.072       & 0.060      & 0.234
                   & 0.064       & 0.087      & 0.168
                   & 0.156       & 0.194
                   & 0.190       & \bf{0.244} \\
      sofa         & 0.457       & 0.450      & 0.548
                   & 0.284       & 0.490      & 0.525
                   & 0.550       & 0.554
                   & 0.560       & \bf{0.569} \\
      table        & 0.159       & 0.116      & 0.305
                   & 0.137       & 0.247      & 0.317
                   & 0.297       & 0.306
                   & 0.305       & \bf{0.320} \\
      \midrule
      Overall      & 0.313       & 0.244      & 0.395
                   & 0.228       & 0.360      & 0.414
                   & 0.419       & 0.400
                   & 0.419       & \bf{0.430} \\
      \bottomrule
    \end{tabular}
  }
  \label{tab:things3d-voxel-reconstruction-iou}
\end{table*}

\begin{table*}[!t]
  \caption{Comparison of single-view 3D object reconstruction on Things3D. We report the mean F-Score@1\% per category and the average F-Score@1\% for all categories. The best number for each category is highlighted in bold.}
  \resizebox{\linewidth}{!} {
    \begin{tabular}{lcccccccccc}
      \toprule
      Category     & 3D-R2N2    & OGN        & Matryoshka
                   & AtlasNet   & Pixel2Mesh & OccNet
                   & IM-Net     & AttSets    
                   & Pix2Vox++/F& Pix2Vox++/A \\
      \midrule
      chair        & 0.166      & 0.096      & 0.231
                   & 0.268      & 0.248      & 0.272
                   & 0.253      & 0.244
                   & 0.240      & \bf{0.273} \\
      display      & 0.136      & 0.126      & 0.164
                   & 0.124      & 0.128      & 0.266
                   & \bf{0.277} & 0.172
                   & 0.150      & 0.163 \\
      lamp         & 0.177      & 0.098      & 0.208
                   & 0.166      & 0.170      & 0.272
                   & 0.248      & 0.229
                   & 0.249      & \bf{0.275} \\
      piano        & 0.012      & 0.006      & 0.127
                   & 0.069      & 0.057      & 0.036
                   & 0.095      & 0.099
                   & 0.108      & \bf{0.136} \\
      sofa         & 0.189      & 0.214      & 0.257
                   & 0.268      & 0.261      & 0.264
                   & 0.259      & 0.253
                   & 0.252      & \bf{0.270} \\
      table        & 0.108      & 0.086      & 0.182
                   & 0.182      & 0.177      & \bf{0.201}
                   & 0.185      & 0.172
                   & 0.198      & 0.200 \\
      \midrule
      Overall      & 0.165      & 0.118      & 0.223
                   & 0.226      & 0.217      & 0.259
                   & 0.244      & 0.231
                   & 0.238      & \bf{0.263} \\
      \bottomrule
    \end{tabular}
  }
  \label{tab:things3d-voxel-reconstruction-fscore}
\end{table*}

\begin{table*}[!t]
  \caption{Comparison of multi-view 3D object reconstruction on Things3D at $32^3$ resolution. We report the mean IoU and F-Score@1\% for all categories.}
  \centering
  \begin{tabularx}{\linewidth}{lYYYYYYYY}
    \toprule
    Methods     & 1 view         & 2 views        & 3 views
                & 4 views        & 5 views        & 6 views
                & 7 views        & 8 views \\
    \midrule
    \multicolumn{9}{l}{\bf{Metric: IoU}} \\
    \midrule
    3D-R2N2     & 0.307          & 0.316          & 0.322
                & 0.325          & 0.329          & 0.331
                & 0.332          & 0.334 \\
    AttSets     & 0.402          & 0.415          & 0.422
                & 0.427          & 0.429          & 0.431
                & 0.433          & 0.434 \\
    Pix2Vox++/F & 0.417          & 0.433          & 0.442
                & 0.447          & 0.451          & 0.454
                & 0.456          & 0.458 \\
    Pix2Vox++/A & \bf{0.428}     & \bf{0.444}     & \bf{0.452}
                & \bf{0.456}     & \bf{0.460}     & \bf{0.462}
                & \bf{0.465}     & \bf{0.467} \\
    \midrule
    \midrule
    \multicolumn{9}{l}{\bf{Metric: F-Score@1\%}} \\
    \midrule
    3D-R2N2     & 0.142          & 0.148          & 0.151
                & 0.153          & 0.156          & 0.157
                & 0.158          & 0.159 \\
    AttSets     & 0.228          & 0.240          & 0.245
                & 0.247          & 0.248          & 0.249
                & 0.250          & 0.250 \\
    Pix2Vox++/F & 0.230          & 0.240          & 0.243
                & 0.246          & 0.248          & 0.249
                & 0.250          & 0.251 \\
    Pix2Vox++/A & \bf{0.260}     & \bf{0.271}     & \bf{0.274}
                & \bf{0.275}     & \bf{0.276}     & \bf{0.277}
                & \bf{0.278}     & \bf{0.279} \\
    \bottomrule
  \end{tabularx}
  \label{tab:things3d-multi-view-reconstruction}
  \vspace{1 mm}
\end{table*}

\begin{table*}
  \caption{The numbers of parameters, inference time, and the corresponding IoUs of Pix2Vox++/A with different backbone models on ShapeNet.}
  \centering
  \begin{tabularx}{\linewidth}{XYYYY}
    \toprule
    \multirow{2}{*}{\shortstack{Pretrained Models}} &
    \multirow{2}{*}{\shortstack{\# Parameters (M)}} &
    \multirow{2}{*}{\shortstack{Inference Time (ms)}} & 
    \multicolumn{2}{c}{IoU} \\
    \noalign{\smallskip} \cline{4-5} \noalign{\smallskip} 
    & & & w/o Pretrained & w/ Pretrained\\
    \midrule
    VGG16        & 97.78       & 11.14      & 0.659 & 0.661 \\
    VGG19        & 98.37       & 11.19      & 0.658 & 0.660 \\
    \midrule
    ResNet50     & \bf{96.31}  & \bf{10.64} & 0.669 & \bf{0.670} \\
    \midrule
    DenseNet101  & 102.85      & 16.78      & 0.668 & 0.669 \\
    DenseNet169  & 109.02      & 18.26      & 0.668 & 0.669 \\
    \bottomrule
  \end{tabularx}
  \label{tab:performance-comparison}
\end{table*}

\subsubsection{Single-view 3D Object Reconstruction}

To evaluate the performance of the proposed methods in dealing with naturalistic images, we compare our methods to several state-of-the-art methods on the Things3D test set.
We fine-tune all competitive methods on Things3D training and crop the input images as required by each method.
To make a fair comparison, all methods are fed with the same input images during testing.
The IoU and F-Score@1\% are reported in Tables \ref{tab:things3d-voxel-reconstruction-iou} and \ref{tab:things3d-voxel-reconstruction-fscore}, respectively.
Both Pix2Vox++/F and Pix2Vox++/A outperform all competitive methods.
Figure \ref{fig:things3d-single-view-reconstruction} shows the qualitative results, which indicate that Pix2Vox++/A has the best ability to recover the 3D shapes from a single natural scene image.

\subsubsection{Multi-view 3D Object Reconstruction}

We also compare the proposed methods with 3D-R2N2 \citep{DBLP:conf/eccv/ChoyXGCS16} and AttSets \citep{DBLP:journals/ijcv/YangSAN19} in reconstructing objects in natural scenes from multi-view images.
As mentioned in Section \ref{sec:things3d}, different objects have different occlusions in different scenes, which leads to different numbers of views for these objects.
To use the same test set in different numbers of views, we only use test samples with no less than eight rendering images in this experiment.
To make a fair comparison, we feed all methods with the same images and crop the images as required by each method during testing.
Table \ref{tab:things3d-multi-view-reconstruction} shows the multi-view reconstruction results on the Things3D test set.
The proposed Pix2Vox++/F and Pix2Vox++/A consistently outperform 3D-R2N2 and AttSets in all numbers of views.
As shown in Figure \ref{fig:things3d-multi-view-reconstruction}, Pix2Vox++/F and Pix2Vox++/A perform better at reconstructing the 3D shape of an object from multiple natural images.
% For example, both 3D-R2N2 and AttSets fail to recover the 3D shape of the piano.
% In contrast, Pix2Vox++/F is able to reconstruct the skeleton of the piano and Pix2Vox++/A can recover the details of the piano.

\section{Analysis and Discussion}

\subsection{Effectiveness of Different Backbone Models}
 
To provide a detailed analysis of different backbone models, we replace ResNet50 in Pix2Vox++/A with other backbone models, including VGG \citep{DBLP:conf/iclr/SimonyanZ14a} and DenseNet \citep{DBLP:conf/cvpr/HuangLMW17}.
We only use partial convolutional layers in backbone models that produce feature maps of size $512 \times 28 \times 28$ to guarantee that the rest of the network architecture of the encoder is the same.
We report the IoU on ShapeNet in Table \ref{tab:performance-comparison}.
Encoders with the pretrained models perform slightly better than those without pretrained models.
Compared to VGG and DenseNet, the encoder with ResNet50 has the best performance in terms of both accuracy and efficiency.

\subsection{Effectiveness of the Refiner}

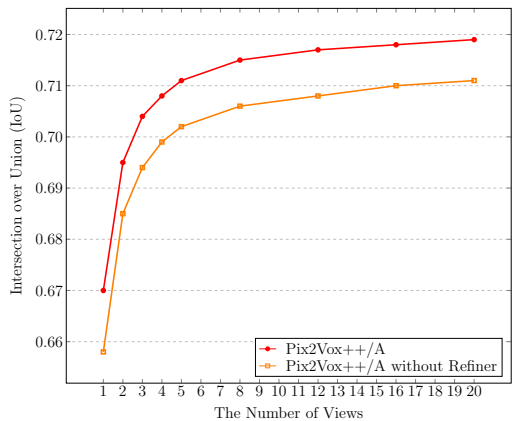
\begin{figure}
  \centering
  \resizebox{.8\linewidth}{!} {
    \begin{tikzpicture}[every plot/.append style={ultra thick}]
      \begin{axis} [
        width = \textwidth,
        xlabel = {The Number of Views},
        xlabel style = {yshift=-0.2cm, font=\fontsize{16}{16}\selectfont},
        ylabel = {Intersection over Union (IoU)},
        ylabel style = {yshift=0.5cm, font=\fontsize{16}{16}\selectfont},
        log basis x = {2},
        xtick = {1, 2, 3, 4, 5, 6, 7, 8, 9, 10, 11, 12, 13, 14, 15, 16, 17, 18, 19, 20},
        xticklabel style = {font=\fontsize{16}{16}\selectfont},
        ytick = {0.63, 0.64, 0.65, 0.66, 0.67, 0.68, 0.69, 0.70, 0.71, 0.72},
        yticklabels = {0.63, 0.64, 0.65, 0.66, 0.67, 0.68, 0.69, 0.70, 0.71, 0.72},
        yticklabel style = {font=\fontsize{16}{16}\selectfont},
        legend style = {at={(0.98, 0.12)}, column sep=0.25cm, font=\fontsize{16}{16}\selectfont},
        legend cell align = left,
        ymajorgrids = true,
        grid style = dashed,
      ]

      \addplot[
        color=red,
        mark=*,
      ]
      coordinates {
        (1, 0.670)(2, 0.695)(3, 0.704)(4, 0.708)(5, 0.711)
        (8, 0.715)(12, 0.717)(16, 0.718)(20, 0.719)
      };
      \addlegendentry{Pix2Vox++/A};

      \addplot[
        color=orange,
        mark=square,
      ]
      coordinates {
        (1, 0.658)(2, 0.685)(3, 0.694)(4, 0.699)(5, 0.702)
        (8, 0.706)(12, 0.708)(16, 0.710)(20, 0.711)
      };
      \addlegendentry{Pix2Vox++/A without Refiner};
      \end{axis}
    \end{tikzpicture}
  }
  \caption{Effectiveness of the refiner and the number of views on the evaluation IoU.}  
  \label{fig:ablation-refiner}
\end{figure}

Pix2Vox++/A uses a refiner to further correct the wrongly recovered parts in the fused 3D volume, which has an IoU of $0.670$ for single-view reconstruction on ShapeNet.
In contrast, IoU decreases to $0.658$ without a refiner.
As shown in Figure \ref{fig:ablation-refiner}, removing the refiner causes considerable degeneration in reconstruction accuracy.

\subsection{Effectiveness of the Camera Parameters}
\label{sec:effectiveness-of-camera-parameters}

Pix2Vox++ recovers the 3D shape of an object without knowing the camera parameters.
It aligns multi-view features with the supervision of ground truth 3D volumes with canonical orientation.
In contrast, LSM \citep{DBLP:conf/nips/KarHM17} aligns multi-view features with the unprojection operation, which requires camera parameters as input.
Table \ref{tab:lsm-multi-view-reconstruction} shows multi-view reconstruction results on ShapeNet compared to LSM.
Experimental results show that LSM significantly outperforms Pix2Vox++/F and Pix2Vox++/A with more than one view, indicating that precise camera parameters help align features of multi-view images better.

To further demonstrate the superior ability of the multi-scale context-aware fusion in multi-view stereo (MVS) systems, we replace the recurrent fusion in LSM with the multi-scale context-aware fusion to fuse features extracted from multiple input images, denoted by LSM-Ctx-Fusion.
As shown in Table \ref{tab:lsm-multi-view-reconstruction}, LSM-Ctx-Fusion outperforms LSM in all numbers of views.

\begin{table}
  \centering
  \caption{Comparison of multi-view 3D object reconstruction on ShapeNet at $32^3$ resolution. We report the mean IoU for all categories. Note that both LSM and LSM-Ctx-Fusion take camera parameters as an additional input. The ShapeNet dataset is provided by LSM \citep{DBLP:conf/nips/KarHM17}.}
  \begin{tabularx}{\linewidth}{lcccc}
    \toprule
    Methods        & 1 view    & 2 views   & 4 views   & 8 views\\
    \midrule
    LSM            & 0.615     & 0.721     & 0.782     & 0.816 \\
    Pix2Vox++/F    & 0.614     & 0.647     & 0.653     & 0.662 \\
    Pix2Vox++/A    & 0.636     & 0.668     & 0.685     & 0.693 \\
    \midrule
    LSM-Ctx-Fusion & \bf{0.639}& \bf{0.739}& \bf{0.806}& \bf{0.838}\\
    \bottomrule
  \end{tabularx}
  \label{tab:lsm-multi-view-reconstruction}
\end{table}

\begin{table}[!t]
  \centering
  \caption{Comparison of multi-view 3D object reconstruction on ShapeNet-Cars at $128^3$ resolution. We report the mean IoU for all categories. The marker $^\dag$ and $^\ddag$ denote the multi-scale context-aware fusion is replaced with the average pooling fusion and context-aware fusion, respectively.}
  \begin{tabularx}{\linewidth}{lcccc}
    \toprule
    Methods        & 1 view    & 2 views   & 4 views   & 8 views\\
    \midrule
    Pix2Vox++/F $^\dag$
                   & 0.803     & 0.804     & 0.805     & 0.806 \\
    Pix2Vox++/F $^\ddag$
                   & 0.803     & 0.784     & 0.778     & 0.768 \\
    Pix2Vox++/F    & 0.803     & 0.813     & 0.815     & 0.819 \\
    \midrule
    Pix2Vox++/A $^\dag$
                   & \bf{0.826}& 0.828     & 0.829     & 0.829 \\
    Pix2Vox++/A $^\ddag$
                   & \bf{0.826}& 0.813     & 0.808     & 0.801 \\
    Pix2Vox++/A    & \bf{0.826}& \bf{0.837}& \bf{0.841}& \bf{0.843}\\
    \bottomrule
  \end{tabularx}
  \label{tab:shapenet-context-aware-fusion-high-resolution}
\end{table}

\begin{figure}
  \resizebox{\linewidth}{!} {
    \includegraphics{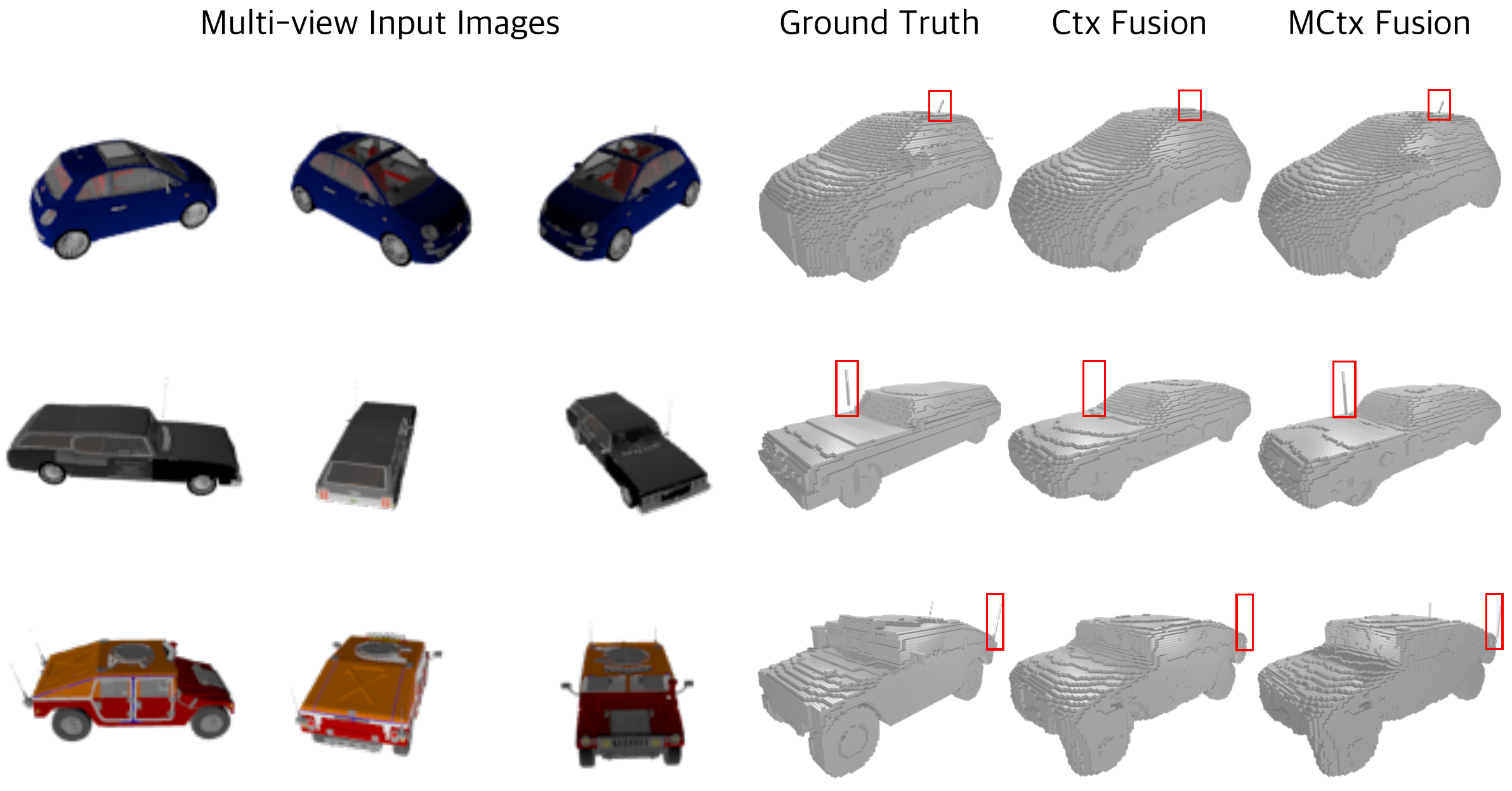}
  }
  \caption{Example of multi-view 3D object reconstruction on ShapeNet-Cars at $128^3$ resolution. ``Ctx Fusion'' and ``MCtx Fusion'' denote the context-aware fusion and the multi-scale context-aware fusion, respectively.}
  \label{fig:multi-scale-context-aware-fusion-comparison}
\end{figure}

\subsection{Comparison with Other Fusion Methods}

To quantitatively evaluate multi-scale context-aware fusion, we replace the multi-scale context-aware fusion in Pix2Vox++/A with the average fusion, context-aware fusion, 3D convolutional LSTM, and attentional aggregation, respectively.

\noindent \textbf{Average Pooling Fusion.}
In the average pooling fusion, the voxel at $(i, j, k)$ among different coarse volumes are averaged.
Specifically, the value of the fused voxel $v^f$ can be calculated as

\begin{equation}
  v^f_{(i, j, k)} = \frac{1}{n}\sum_{r = 1}^n v_{(i, j, k)}^r
\end{equation}
As shown in Table \ref{tab:shapenet-context-aware-fusion}, replacing the multi-scale context-aware fusion with the average fusion in  Pix2Vox++/F and Pix2Vox++/A causes degeneration in reconstruction results.

\noindent \textbf{Context-aware Fusion.}
We also compare the multi-scale context-aware fusion with the context-aware fusion in the preliminary version \citep{DBLP:conf/iccv/XieHXSS19}, denoted by Pix2Vox++/F$^\ddag$ and Pix2Vox++/A$^\ddag$, respectively.
As shown in Table \ref{tab:shapenet-context-aware-fusion}, IoU has an approximately 1\% increase in the multi-scale context-aware fusion compared to the context-aware fusion.
Accurately reconstructing a 3D volume at high resolution is challenging due to increasing voxels reflecting object details.
Both the average fusion and context-aware fusion may not fully exploit the features of multi-view images.
In contrast, the proposed multi-scale context-aware fusion preserves object details by concatenating feature maps with different scales.
As shown in Table \ref{tab:shapenet-context-aware-fusion-high-resolution}, the multi-scale context-aware fusion outperforms the average fusion and the context-aware fusion in reconstructing high-resolution 3D volumes.
As illustrated in Figure \ref{fig:multi-scale-context-aware-fusion-comparison}, the multi-scale context-aware fusion recovers better details than the context-aware fusion.

\begin{table}[!t]
  \centering
  \caption{Comparison of single-view 3D object reconstruction on the ShapeNetCore dataset. We report the mean IoU for all categories in both object-centered and viewer-centered coordinates. Note that ``Unseen'' denotes no instances from the categories are seen during training.}
  \begin{tabularx}{\linewidth}{lYYYY}
    \toprule
    \multirow{2}{*}{Methods} & 
    \multicolumn{2}{c}{Object-centered} &
    \multicolumn{2}{c}{Viewer-centered} \\
    \cline{2-5}
                & Seen       & Unseen     & Seen       & Unseen \\
    \midrule
    OGN         & 0.593      & 0.154      & 0.404      & 0.267 \\
    Matryoshka  & 0.634      & 0.187      & 0.427      & 0.299 \\
    Pix2Vox++/F & 0.632      & 0.216      & 0.449      & 0.324 \\
    Pix2Vox++/A & \bf{0.688} & \bf{0.241} & \bf{0.485} & \bf{0.386} \\
    \bottomrule
  \end{tabularx}
  \label{tab:object-viewer-centered-coordinates}
\end{table}

\begin{table*}[!t]
  \caption{Comparison of multi-view 3D object reconstruction on ShapeNet at $32^3$ resolution. We report the mean IoU per category and the average IoU for all categories. The marker $^\dag$ and $^\ddag$ denote the multi-scale context-aware fusion is replaced with the average pooling fusion and context-aware fusion, respectively.}
  \centering
  \begin{tabularx}{\linewidth}{lYYYYYYccc}
    \toprule
    
    Methods     & 1 view         & 2 views        & 3 views
                & 4 views        & 5 views        & 8 views
                & 12 views       & 16 views       & 20 views \\
    \midrule
    Pix2Vox++/F \hspace{.7mm}$^\dag$
                & 0.645          & 0.655          & 0.664
                & 0.668          & 0.670          & 0.672 
                & 0.673          & 0.674          & 0.675 \\
    Pix2Vox++/F \hspace{.7mm}$^\ddag$
                & 0.645          & 0.663          & 0.673
                & 0.676          & 0.680          & 0.683 
                & 0.686          & 0.687          & 0.688 \\
    Pix2Vox++/F  & 0.645         & 0.669          & 0.678
                & 0.682          & 0.685          & 0.690
                & 0.692          & 0.693          & 0.694 \\
    \midrule
    \midrule
    Pix2Vox++/A \hspace{.1mm}$^\dag$
                & \bf{0.670}     & 0.680          & 0.690
                & 0.695          & 0.699          & 0.703
                & 0.704          & 0.705          & 0.706 \\
    Pix2Vox++/A \hspace{.1mm}$^\ddag$
                & \bf{0.670}     & 0.690          & 0.699
                & 0.702          & 0.706          & 0.710
                & 0.712          & 0.713          & 0.714 \\
    Pix2Vox++/A-R2N2
                & 0.663          & 0.672          & 0.680
                & 0.684          & 0.686          & 0.688
                & 0.689          & 0.689          & 0.690 \\
    Pix2Vox++/A-AttSets
                & 0.638          & 0.675          & 0.689
                & 0.696          & 0.701          & 0.707
                & 0.710          & 0.713          & 0.713 \\
    \midrule
    Pix2Vox++/A & \bf{0.670}     & \bf{0.695}     & \bf{0.704}
                & \bf{0.708}     & \bf{0.711}     & \bf{0.715}
                & \bf{0.717}     & \bf{0.718}     & \bf{0.719} \\
    \bottomrule
  \end{tabularx}
  \label{tab:shapenet-context-aware-fusion}
\end{table*}

\begin{table*}
  \caption{The numbers of parameters, memory footprint, and inference time on the ShapeNet dataset. Note that the memory is measured in backward computation of single-view reconstruction with a batch size of 1. The voxel reconstruction methods, including 3D-R2N2, AttSets, Pix2Vox++/F, and Pix2Vox++/A, output 3D volumes at $32^3$ resolution.}
  \centering
  \begin{tabularx}{\linewidth}{l|cccc|cccc}
  \toprule
  Methods               & AtlasNet    & Pixel2Mesh
                        & OccNet      & IM-Net
                        & 3D-R2N2     & AttSets
                        & Pix2Vox++/F & Pix2Vox++/A \\
  \midrule
  \#Parameters (M)      & 45.06       & 21.36
                        & 13.43       & 55.45
                        & 35.97       & 17.71
                        & 4.83        & 96.31 \\
  Memory (MB)           & 1293        & 1289
                        & 955         & 3935
                        & 1407        & 3911
                        & 647         & 2411 \\
  \midrule
  \midrule
  \multicolumn{8}{l}{\bf{Inference Time (ms)}} \\
  \midrule
  1 view                & 38.47       & 60.78
                        & 1261        & 10886
                        & 78.86       & 26.32
                        & 9.93        & 10.64 \\
  2 views               & N/A         & N/A
                        & N/A         & N/A
                        & 112.27      & 47.62 
                        & 13.55       & 17.51 \\
  4 views               & N/A         & N/A
                        & N/A         & N/A
                        & 116.68      & 52.63
                        & 23.72       & 29.88 \\
  8 views               & N/A         & N/A
                        & N/A         & N/A
                        & 122.04      & 58.83
                        & 39.02       & 56.52 \\
  \bottomrule
  \end{tabularx}
  \label{tab:space-time-complexity}
\end{table*}

\noindent \textbf{3D Convolutional LSTM.}
To further compare with RNN-based fusion, we remove the multi-scale context-aware fusion from Pix2Vox++/A and add a 3D convolutional LSTM \citep{DBLP:conf/eccv/ChoyXGCS16} after the encoder.
To fit the 3D convolutional LSTM input, we add an additional fully connected layer with a dimension of $1024$ before it.
The resulting method is named Pix2Vox++/A-R2N2.
As shown in Table \ref{tab:shapenet-context-aware-fusion}, both Pix2Vox++/A and Pix2Vox++/A$^\dag$ consistently outperform Pix2Vox++/A-R2N2 in all numbers of views. 

\noindent \textbf{Attentional Aggregation.}
To demonstrate the superior reconstruction ability over the attentional aggregation \citep{DBLP:journals/ijcv/YangSAN19}, we remove the multi-scale context-aware fusion from Pix2Vox++/A and add an attentional aggregation module after the encoder, denoted by Pix2Vox++/A-AttSets.
Experimental results in Table \ref{tab:shapenet-context-aware-fusion} show that Pix2Vox++/A outperforms Pix2Vox++/A-AttSets in all numbers of views.

\subsection{Viewer-centered vs. Object-centered Coordinates}

As mentioned in Section \ref{sec:effectiveness-of-camera-parameters}, Pix2Vox++ relies on the object-centered coordinates to align multi-view features.
However, object-centered coordinates encourage the network to memorize observed meshes, which may lead to poor generalization abilities \citep{DBLP:conf/cvpr/ShinFH18}.

To evaluate the generalization capability of the proposed methods, we compare the performance of reconstructing 3D objects from ``seen'' and ``unseen'' categories in both viewer-centered and object-centered coordinates.
For object-centered prediction, different views of the same object should produce the same 3D shape.
In viewer-centered coordinates, the reconstructed 3D object should be oriented  according to the input viewpoint, so different views of the same object correspond to different 3D shapes.

In this experiment, we use Blender to render objects in 57 categories of ShapeNetCore \citep{DBLP:journals/corr/ChangFGHHLSSSSX15} from 24 random views for each object. 
When reconstructing 3D objects from ``unseen'' categories, all pretrained models have never ``seen'' either the objects in these categories or the labels of the objects before.
More specifically, all methods are pretrained on the 13 major categories of ShapeNet and tested on the remaining 44 categories of ShapeNetCore with the same input images.
As shown in Table \ref{tab:object-viewer-centered-coordinates}, 3D shapes produced by object-centered models outperform those produced by viewer-centered models for objects from ``seen'' categories, suggesting that reconstructing viewer-centered 3D shapes of objects is more challenging.
For ``unseen'' categories, viewer-centered models perform better than object-centered models, suggesting that viewer-centered reconstruction improves the generalization ability to reconstruct 3D shapes from ``unseen'' categories.
Compared to OGN \citep{DBLP:conf/iccv/TatarchenkoDB17} and Matryoshka Networks \citep{DBLP:conf/cvpr/Richter018}, Pix2Vox++/F and Pix2Vox++/A perform better at reconstructing 3D shapes in both object-centered and viewer-centered coordinates.

\subsection{Space and Time Complexity}

To test the space and time complexity of our methods, we compare them with several state-of-the-art methods in terms of number of parameters, memory usage, and inference time.
Table \ref{tab:space-time-complexity} presents the comparison results in single-view and multi-view reconstruction, where the voxel reconstruction methods are at $32^3$ resolution.
Table \ref{tab:space-time-complexity-high-resolution} provides a comparison of the single-view reconstruction results at $128^3$ resolution.

Running times are obtained on the same PC with a single NVIDIA GTX 1080 Ti GPU.
For more precise timing, we exclude reading and writing time when evaluating inference time.
For multi-view reconstruction, both Pix2Vox++/F and Pix2Vox++/A outperform 3D-R2N2 and AttSets in inference time and training time.
Both Pix2Vox++/F and Pix2Vox++/A are approximately seven times faster in forward inference than 3D-R2N2 for single-view reconstruction at $32^3$ resolution.
Although the proposed methods outperform OGN \citep{DBLP:conf/iccv/TatarchenkoDB17} and Matryoshka Networks \citep{DBLP:conf/cvpr/Richter018} in reconstructing high-resolution 3D volumes, memory requirements scale dramatically with the resolution of 3D volumes because our methods do not use efficient data representations.

\begin{table}
  \caption{The numbers of parameters, memory footprint, and inference time at $128^3$ resolution on the ShapeNet dataset. P2V/F and P2V/A denote Pix2Vox++/F and Pix2Vox++/A for $128^3$ resolution. ``Inf. Time'' stands for ``Inference Time'' for single-view reconstruction with a batch size of 1. Note that the memory is measured in backward computation of single-view reconstruction with a batch size of 1.}
  \centering
  \begin{tabularx}{\linewidth}{lcccc}
  \toprule
  Methods         & OGN         & Matryoshka
                  & P2V/F       & P2V/A \\
  \midrule
  \#Params (M)    & 12.46       & 45.66
                  & 5.02        & 96.57 \\
  Memory (MB)     & 861         & 1593
                  & 2227        & 3997 \\
  Inf. Time (ms)  & 52.87       & 11.92
                  & 37.10       & 51.96 \\
  \bottomrule
  \end{tabularx}
  \label{tab:space-time-complexity-high-resolution}
\end{table}

\section{Conclusion}

In this paper, we propose a unified framework for both single-view and multi-view 3D reconstruction, named Pix2Vox++.
Compared with existing methods that directly fuse the features from multi-view images, the proposed framework fuses the 3D volumes reconstructed from input images, which better preserves multi-view spatial constraints.
In addition, we construct the first large-scale naturalistic dataset for multi-view 3D object reconstruction, named {\it Things3D}, containing 1.68M images of 280K objects collected from over 39K indoor scenes.
Quantitative and qualitative evaluation for both single-view and multi-view reconstruction on the ShapeNet, Pix3D, and Things3D benchmarks shows that the proposed methods perform favorably against state-of-the-art methods.
The proposed methods are also computationally efficient, about seven times faster than 3D-R2N2 in terms of inference time in single-view reconstruction.

\begin{acknowledgements}
This work was supported in part by the National Natural Science Foundation of China under Pro-ject (Nos. 61772158, 61702136 and 61872112), in part by National Key Research and Development Program of China (Nos. 2018YFC0806802 and 2018YFC0832105), and in part by Self-Planned Task (No. SKLRS202002D) from the State Key Laboratory of Robotics and System (HIT).
We would like to thank anonymous reviewers for their valuable feedback during this research.
\end{acknowledgements}


\bibliographystyle{spbasic}       % basic style, author-year citations
\bibliography{references}   % name your BibTeX data base

\end{document}